%% file: iclr2027_conference_arxiv.tex
\documentclass{article} % For LaTeX2e
\usepackage{iclr2027_conference,times}
\usepackage{booktabs}
\usepackage{graphicx}
\usepackage{flafter}
\usepackage{float}
\usepackage{wrapfig}
\usepackage{enumitem}
\usepackage{listings}
\usepackage{amsmath} 

\input{table/benchmark-table-style.tex}

\renewcommand{\BTiconpath}{table/model-icons}

\iclrfinalcopy

\input{math_commands.tex}

\usepackage{hyperref}
\usepackage{url}

\title{
ExeWorldBench: Benchmarking Code-Based Executable 3D World Modeling
}

\author{
Xinge Peng$^{1}$,
Zhen Fang$^{1}$,
Ruibing Wang$^{1}$,
Yongxuan Lv$^{1}$,
Haodong Li$^{2}$,
Guohui Zhang$^{1}$,\\
Xiaoxiao Ma$^{1}$,
Yiting Lu$^{1}$,
Lin Chen$^{1}$,
Zehui Chen$^{1}$,
Jie Huang$^{1}$,
Feng Zhao$^{1}$
\\[2mm]
$^{1}$University of Science and Technology of China \\
$^{2}$The Chinese University of Hong Kong \\
\texttt{xg.pengv@gmail.com}
}

\begin{document}

\maketitle

\vspace{-5mm}

\begin{center}
\url{https://helen1p.github.io/ExeWorldBench/}
\end{center}

% \vspace{-1mm}
\begin{abstract}
\vspace{-3mm}
Coding agents are increasingly evolving beyond reproducing \textit{what a world looks like} toward constructing executable worlds that capture \textit{how a world works}. This shift calls for \textbf{executable world modeling}, which integrates two core capabilities: \textbf{world understanding}, reasoning about the underlying structural, mechanical, and causal principles of a world, and \textbf{executable realization}, translating that understanding into a code-based system whose behavior emerges from physical interactions. However, existing benchmarks provide only fragmented evaluations centered on \textit{physical reasoning}, \textit{visual fidelity}, or \textit{behavioral correctness}, leaving the integration of these capabilities insufficiently assessed. To bridge this gap, we introduce \textbf{ExeWorldBench}, to our knowledge the first benchmark to comprehensively evaluate world understanding and executable realization through code-based 3D world construction. ExeWorldBench comprises three progressively challenging tasks: \textbf{Mechanism Executability}, which evaluates the realization of individual mechanisms from visual observations; \textbf{Compositional Executability}, which extends to multi-stage systems with spatial, mechanical, and causal dependencies; and \textbf{Intent-Driven Executable World Synthesis}, which requires constructing executable worlds from high-level goals and physical requirements. Extensive experiments on frontier models reveal substantial limitations in executable world modeling, offering insights into the development of more physically grounded and causally coherent world construction systems.

\end{abstract}
\vspace{-3mm}
\section{Introduction}
\vspace{-3mm}

Coding agents are evolving beyond their role as general-purpose programming tools toward constructing worlds. 
Conventional visual-to-code generation~\citep{zhang2026widget2code,zhang2026visual,deng2026visrefiner,jiang2025screencoder} primarily focuses on reproducing \textit{what a world looks like}, translating visual observations into code that reconstructs the appearance, layout, and geometric structure of the reference. 
However, constructing an executable world goes beyond visual fidelity, requiring not only a faithful representation of its structure but also the physical mechanisms and causal interactions that govern \textit{how the world works}. 
This raises a fundamental question: \textbf{Can current models construct not only what a world looks like, but also how it works?}

\vspace{-1.0mm}

Constructing such worlds requires two tightly coupled capabilities. 
The first is \textbf{world understanding}, which involves reasoning about the latent structural, mechanical, and causal principles that govern how a world is organized and evolves. 
The second is \textbf{executable realization}, which requires translating this understanding into an executable code-based system. 
Crucially, these capabilities cannot be considered in isolation: \textit{understanding how a world works does not guarantee that it can be faithfully realized, while generating runnable code does not necessarily indicate that the underlying world has been correctly understood. }

Existing benchmarks provide only partial evidence of this integrated capability. 
On one side, \textbf{reasoning-centric} benchmarks~\citep{chow2025physbench,zhou2026pai,li2024mvp,peng2026physvista} typically evaluate world understanding through physical reasoning tasks formulated as question answering over images or videos, testing whether models can infer physical properties, relations, or outcomes without requiring them to realize such understanding in an executable world. 
Such evaluations can reveal whether a model understands aspects of physical behavior, but not whether this understanding can be effectively integrated into executable construction. 
On the other side, \textbf{construction-centric} benchmarks~\citep{gao20263dcodebench,noor2026nova3d,zhang2026benchcad,wu2025plot2code,si2025design2code} primarily evaluate whether models can translate visual or textual specifications into valid code that reproduces the intended visual or geometric structure, with evaluation largely centered on appearance, geometry, and static structural fidelity. 
They therefore provide limited evidence about whether the structure and principles inferred from a world are faithfully translated into its executable realization. 

\textbf{Behavior-centric} benchmarks~\citep{xie2026physcodebench,wang2026simuscene,lu2026worldcoder,liang2026visphyworld,kovavcivc2026mpmworlds} often evaluate \textit{behavioral correctness}, such as whether objects follow expected trajectories, reach specified states, or reproduce observed or intended dynamics. However, these behaviors are often relatively simple or isolated, such as falling, sliding, or single-object motion, whereas executable world modeling may involve coordinated interactions among multiple components, introducing richer spatial, mechanical, and causal dependencies. 
Additionally, while behavioral criteria provide useful evidence of dynamic correctness, they offer only a partial assessment of the model's causal understanding. 
In simple or unidirectional motions, plausible trajectories can sometimes be reproduced through interpolation between observed states or predefined animations, without faithfully implementing the underlying physical mechanisms. 
Thus, reproducing \textit{how a world moves} does not necessarily demonstrate that the model can construct an executable world that reflects \textit{how the world works}.

Beyond the evaluation target itself, some existing benchmarks operate in simplified physical or representational domains~\citep{kovavcivc2026mpmworlds,wang2026simuscene,lu2026worldcoder}. 
In particular, 2D simulations and browser-native environments abstract away several challenges that become central in executable 3D worlds, including spatial arrangement, depth-dependent contacts, collision geometry, mechanical assembly, and joint alignment. 
Therefore, a comprehensive evaluation of executable world modeling should go beyond isolated behaviors and simplified environments to assess how models integrate the understanding of complex physical worlds with executable realization.

To address these limitations, we introduce \textbf{ExeWorldBench}, a benchmark for executable 3D world modeling that progressively increases both world complexity and model autonomy. It systematically probes the transition from observation-grounded realization to autonomous world construction, gradually reducing the guidance available for determining how a world should be built. 
To operationalize this progression, ExeWorldBench comprises three complementary tasks: \textbf{Mechanism Executability} evaluates the reconstruction of individual mechanical mechanisms from visual observations; \textbf{Compositional Executability} extends this to multi-stage mechanical systems requiring coordinated spatial, mechanical, and causal interactions; and \textbf{Intent-Driven Executable World Synthesis} moves beyond visual references to evaluate the construction of executable worlds from high-level textual specifications. 
The final task further includes two settings: \textit{Component-Constrained Synthesis}, which specifies the required physical components while leaving their arrangement and causal connections to the model, and \textit{Goal-Constrained Synthesis}, which provides higher-level goals and physical requirements while requiring the model to determine the underlying mechanical realization. 
Together, these tasks bridge the gap between understanding how a world works and constructing a world that works, probing whether models can faithfully translate physical world understanding into executable realization.

Our contributions are threefold:
\vspace{-2.5mm}
\begin{itemize}[leftmargin=1.5em]
\item We formulate \textbf{executable 3D world modeling} as the integration of world understanding and executable realization, requiring models to reason about the geometric, mechanical, and causal organization of a world and translate that understanding into an executable code-based system.

\item We introduce \textbf{ExeWorldBench}, to our knowledge the first benchmark to comprehensively evaluate how models translate their understanding of \textit{how a world works} into executable realization, through three progressively challenging tasks: \textbf{Mechanism Executability}, \textbf{Compositional Executability}, and \textbf{Intent-Driven Executable World Synthesis}.

\item We benchmark frontier models and reveal substantial gaps between code executability and physically coherent world construction, highlighting the need for tighter integration of world understanding and executable realization.
\end{itemize}

\vspace{-3mm}
\section{Related Work}
\vspace{-3mm}

\noindent\textbf{Approaches to Code-Based World Modeling.}
Recent work increasingly treats code as an executable representation of dynamic worlds.
One line of work focuses on \textit{specification-guided world construction}, which takes explicit \textit{textual descriptions} as input and turns them into executable world that exhibit the specified behaviors, as explored by Code2Worlds~\citep{zhang2026code2worlds}, WorldClaw~\citep{guo2026worldclaw}, coding-agent-based world simulators~\citep{wang2026coding}, and interactive environment generation~\citep{kang2026simworld}. 
Related work further studies executable programs as persistent world representations, including Code-as-World~\citep{wang2026code} and Code World Model~\citep{chen2026code}.
A complementary line, \textit{observation-based world discovery}~\citep{cherian2024llmphy,yang2026physmind}, which begins with \textit{visual observations} and seeks to infer the hidden physical properties, dynamics, and structural organization that explain the world. 
Overall, existing methods only achieved fragmented world modeling capability. 
Executable world modeling goes beyond either paradigm alone, requiring both comprehensive \textbf{world understanding} and the ability to \textbf{translate that understanding into an executable code-based system}, rather than merely implementing prescribed behaviors. 

\vspace{-1mm}

\noindent\textbf{Benchmarks for Code-Based World Modeling.}
Existing benchmarks cover several aspects of code-based physical world construction.
PhysCodeBench~\citep{xie2026physcodebench} and SimuScene~\citep{wang2026simuscene} evaluate executable realization from textual physical specifications, while WorldCoder-Bench~\citep{lu2026worldcoder} extends this to browser-native 3D environments with spatial and behavioral constraints.
3DCodeBench~\citep{gao20263dcodebench} and BenchCAD~\citep{zhang2026benchcad} focus on executable geometric construction from text or visual input.
VisPhyWorld~\citep{liang2026visphyworld} and MPMWorlds~\citep{kovavcivc2026mpmworlds} instead evaluate reconstruction or prediction of physical dynamics from visual observations, while PACE-Bench~\citep{zhan2026pace} studies simulation-code adaptation under changing physical conditions.
Despite these formulations, existing benchmarks mainly emphasize behavioral reproduction, geometric fidelity, parameter recovery, or dynamics adaptation, providing limited evaluation of coherent 3D mechanisms whose geometry, constraints, contacts, and physical interactions jointly produce complex causal behavior.

\begin{figure*}[t]
\vspace{-10mm}
\includegraphics[width=1.0\linewidth]{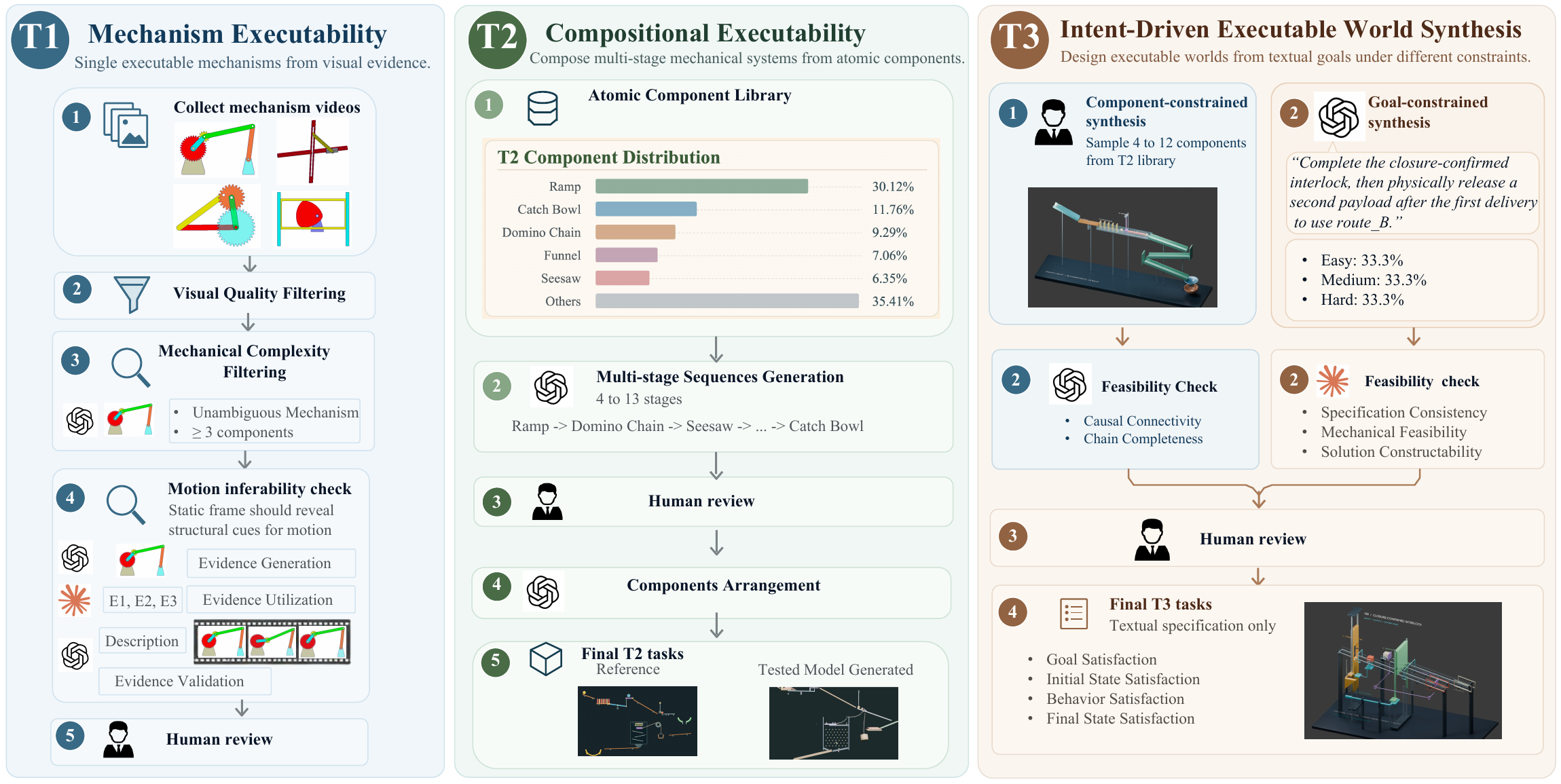}
  \centering
  \vspace{-1.3em}
  \caption{\textbf{Data construction pipeline of ExeWorldBench.}
T1 filters real mechanical mechanisms, T2 composes multi-stage systems from atomic components, and T3 constructs component- and goal-constrained synthesis tasks with feasibility verification.}
\vspace{-1.4em}
\label{fig:data}
\end{figure*}

\vspace{-3mm}
\section{ExeWorldBench}
\vspace{-3mm}
\subsection{Task Formulation}
\vspace{-3mm}

ExeWorldBench evaluates executable 3D world modeling through three progressively challenging tasks that probe how models translate physical understanding into executable construction.

\textbf{T1: Mechanism Executability} evaluates whether a model can infer and realize a complex mechanical mechanism from visual observations. Given a reference image, the model must generate a Blender program that constructs the mechanism and executes its complete motion. 

This task is challenging in two respects:
(i) \textbf{motion complexity}: rather than following simple object trajectories, the target motion emerges from coupled interactions among multiple mechanical components;
(ii) \textbf{limited motion observation}: the input primarily specifies the mechanism structure through a single reference image, rather than exposing its dynamics through sampled motion frames. 
The model must therefore infer the complete motion from structural evidence, including the underlying constraints, contacts, and motion propagation, rather than following or extrapolating an observed motion sequence.

\textbf{T2: Compositional Executability} extends T1 from a single mechanism to a multi-stage mechanical system, requiring multiple mechanisms to form a complete executable causal chain.

The task introduces two key challenges:
(i) \textbf{reasoning over both individual mechanisms and causal interactions}: the model must understand the behavior of each mechanism while also reasoning about how motion and effects propagate across multiple stages; 
(ii) \textbf{spatial arrangement requirements}: the model must determine appropriate component positions, orientations, and distances so that each stage can physically connect to and trigger the next.

\textbf{T3: Intent-Driven Executable World Synthesis} removes direct visual references and requires the model to construct an executable world from high-level textual specifications. 

We consider two settings.
\textbf{T3-1: Component-Constrained Synthesis} provides a set of required physical components and requires the model to organize them into a complete executable system.
\textbf{T3-2: Goal-Constrained Synthesis} instead specifies higher-level task requirements, including the goal, required objects, state and event definitions, initial state, physical rules and dependencies, and final state, while leaving the underlying mechanical realization to the model. 
% In both settings, the model must determine the geometry, spatial layout, and mechanical realization of the system. 

Unlike T1 and T2, where the target scene is largely specified by the visual input, T3 requires the model to determine the scene configuration and mechanical realization itself. 
This demands more open-ended reasoning over how high-level physical intent can be translated into a coherent 3D mechanism whose behavior emerges through executable physical interactions.

\vspace{-3mm}
\subsection{Data Construction}
\vspace{-3mm}
We construct the three tasks through distinct data construction pipelines, as illustrated in Fig.~\ref{fig:data}. 
\vspace{-3mm}
\subsubsection{T1: Mechanism Executability}
\vspace{-3mm}
\textbf{Data Construction.}
We collect publicly available animations illustrating mechanical principles and dynamic mechanical systems. 
Each sample provides a clear view of the mechanism together with its complete motion cycle. 
We retain samples that satisfy three criteria:
(i) the visual quality is sufficiently high to identify key components and their spatial relationships, without severe distortions (\textit{e.g.}, blur, JPEG compression);
(ii) the mechanism involves meaningful mechanical interactions among multiple moving components;
and (iii) the complete motion cycle can be reasonably inferred from the visible static structure. 

\textbf{Data Verification.}
We first use a visual quality assessment model to filter visually degraded samples. 
GPT-6 Astra is then used for mechanical complexity filtering, which filters cases with severe occlusion or ambiguous component boundaries, or fewer than three counts mechanically active components.

To assess \textbf{motion inferability}, we follow a generate--infer--validate--verify pipeline:
 \vspace{-0.5mm}
 
(i) \textbf{Evidence generation.}
The static reference image together with the full sequence of motion reference frames is provided to GPT-6 Astra. 
Over three interactive rounds, GPT-6 Astra extracts three pieces of visual evidence from the static image that can be used to infer the target motion observed in the reference sequence. 
The extracted evidence is restricted to information directly identifiable from the static image and cannot explicitly describe the motion itself. 

(ii) \textbf{Evidence utilization.}
The static reference image and generated evidence are provided to Opus 5, a model of comparable capability, without access to the motion reference sequence. 
A complete description of the mechanism motion is then derived from the available evidence, including major motion patterns, component relationships, and mechanical interactions. 

(iii) \textbf{Evidence validation.}
Given the generated motion description and the motion reference sequence, GPT-6 Astra performs a cross-modal motion consistency check.
Only samples whose described dynamics are consistent with the reference motion are retained, thereby validating the reliability of the extracted evidence and further supporting the motion inferability of the static input. 

(iv) \textbf{Human verification.}
Finally, the retained samples are manually reviewed by five human annotators, and cases with low motion inferability or ambiguous structural evidence are removed. 

\vspace{-3mm}
\subsubsection{T2: Compositional Executability}
\vspace{-3mm}

\textbf{Data Construction.}
We first construct a code library of atomic mechanical components with predefined motion patterns. 
Only components with visually identifiable and unambiguous motion, and with behavior reflecting meaningful physical or mechanical principles rather than purely decorative animation, are retained as reusable building blocks. 
GPT-6 Astra is then used to generate ordered textual sequences of components for multi-step mechanical systems ranging from 4 to 13 steps, together with the intended transitions between consecutive stages. 

\textbf{Data Verification and Reference Generation.}
Each generated sequence is manually reviewed to ensure that the component order and stage-to-stage transitions are physically coherent. 
For each retained sequence, GPT-6 Astra (Codex CLI) is used to assemble the corresponding component implementations from our code library according to the given ordered component sequence.
Its role is primarily to organize these predefined component implementations and adjust their spatial layout to produce a continuous reference animation.

\vspace{-3mm}
\subsubsection{T3: Intent-Driven Executable World Synthesis}
\vspace{-2mm}
\textbf{Data Construction.}
For \textbf{Component-Constrained Synthesis}, we reuse the atomic component library from T2 and randomly sample 4 to 12 components to form tasks with 4 to 12 stages. 
For \textbf{Goal-Constrained Synthesis}, we construct tasks at three difficulty levels: easy, medium, and hard.
GPT-6 Astra is prompted with the corresponding difficulty criteria and a predefined schema to generate complete task specifications, including the goal, required objects, state and event definitions, initial state, physical rules and dependencies, and final state.

\textbf{Data Verification.}
For \textbf{Component-Constrained Synthesis}, each sampled component set, together with the predefined motion pattern of each component, is provided to GPT-6 Astra for feasibility checking. 
The model verifies whether
(i) the components admit a coherent ordering with physically plausible stage-to-stage causal connections, and
(ii) all sampled components can participate in a single complete execution chain without unused or isolated stages. 
Only feasible component sets are retained. 
For \textbf{Goal-Constrained Synthesis}, each generated task is validated by Opus 5 along three aspects:
(i) whether the specification is internally consistent, clear, and unambiguous;
(ii) whether the task is mechanically feasible and solvable; and
(iii) whether a valid reference solution can be constructed.
Tasks that fail any of these checks are discarded.
Finally, the retained tasks are manually reviewed to further verify specification consistency, mechanical feasibility, and solution validity.

\vspace{-1mm}
\section{Evaluation}
\vspace{-3mm}
We evaluate a diverse set of frontier models on all three tasks, covering both harness-based and API-based evaluation settings.
\vspace{-3mm}
\subsection{Metrics}
\vspace{-3mm}
We evaluate the generated executable worlds along four dimensions, using Opus 5.5 as the VLM judge.
By default, we report \emph{penalized} metrics over all examples, including cases where code execution fails.
For completeness, we also report \emph{conditioned} metrics computed only over code-executable cases in Appendix.~\ref{sec:app1}. 
Subscripts 1 and 2 of each metric denote results before and after execution-error repair, respectively.

\noindent\textbf{Visual Structure ($V$)} evaluates the \emph{static} quality and physical validity of generated worlds across all tasks.
It considers component geometry, spatial arrangement, mechanical assembly, and inter-component relationships, while penalizing defects such as interpenetration, disconnection, implausible configurations, and severe rendering artifacts.
Scores range from 0 to 10.
For reference-guided tasks, the judge considers the provided visual references, while synthesis tasks are evaluated against their task specifications.

\noindent\textbf{Mechanical Motion ($M$)} is used in T1 to evaluate the \emph{dynamic behavior of individual mechanical components}.
It measures whether components exhibit the intended motion type, direction, range, and temporal evolution throughout execution.
Scores range from 0 to 10, based on comparison between sampled generated frames and the complete reference frames.

\noindent\textbf{Executable Causal Mechanism (ECM)} evaluates the \emph{dynamic execution and causal interactions} of multi-stage systems in T2 and T3.
Each execution is decomposed into causal segments, where the judge assesses both component behavior and the correctness of each stage-to-stage causal handoff.
Each segment is scored from 0 to 10, and ECM is averaged over all segments.

\noindent\textbf{Goal Satisfaction ($G$)} is used in T3-2 to evaluate task-specific requirements on the initial state ($I$), intermediate behavior ($B$), and final state ($F$).
Each condition is scored binary: 1 only if fully satisfied, and 0 otherwise.
The judge assesses these conditions from the sampled generated frames together with the task-specific object, event, and state definitions.

Given $N$ evaluation samples, overall Goal Satisfaction is:
\vspace{-1mm}
\begin{equation}
\text{Goal Sat.}
=
\frac{1}{N}
\sum_{i=1}^{N}
\frac{P_i^{I}+P_i^{B}+P_i^{F}}
{N_i^{I}+N_i^{B}+N_i^{F}},
\label{eq:goal_satisfaction}
\end{equation}
where $P_i^{c}$ and $N_i^{c}$ denote the numbers of satisfied and total conditions for category $c$ in sample $i$.

We also report category-wise satisfaction:
\begin{equation}
\text{Sat.}_{c}
=
\frac{1}{N}
\sum_{i=1}^{N}
\frac{P_i^{c}}
{N_i^{c}},
\qquad c \in \{I,B,F\}.
\label{eq:category_satisfaction}
\end{equation}

\vspace{-8mm}
\subsection{Code Executability}
\vspace{-3mm}
Code executability is reported in the first two columns of Tab.~\ref{tab:t1}, Tab.~\ref{tab:t2}, Tab.~\ref{tab:t3-1}, and Tab.~\ref{tab:t3-2}.
The first line shows executability from initial single-turn generation.
GPT-6 Astra~\citep{openai2026gpt6astra} and Fable 5.1~\citep{anthropic2026fable51} achieve relatively high success among API-based models, while the harness-based models reach nearly 100\% executability on most tasks, indicating greater robustness in producing runnable Blender programs.

To separate executable world modeling from incidental implementation failures, we use GPT-6 Astra (Codex CLI)~\citep{openai2026gpt6astra,openai_codex_cli} to repair API-based outputs by correcting only technical execution errors without changing the intended scene structure or behavior.
The second line reports executability after repair, which improves substantially across models, showing that many initial failures stem from low-level implementation issues.

\textbf{High code executability does not imply strong executable-world modeling performance.}
After repair, Doubao-Seed-2.1-Pro~\citep{bytedanceseed2026seed21}, DeepSeek-V4.1-Flash~\citep{deepseekai2026deepseekv41flash}, Qwen3.8-Max~\citep{alibaba2026qwen38max}, Kimi K3~\citep{kimiteam2026kimik3}, and GLM-5.3-Flash~\citep{zai2026glm53flash} all achieve relatively high executability, while the other metrics improve only modestly. 
In other words, the gain in executability is substantially larger than the gains in the corresponding world modeling metrics.
This gap shows that producing runnable Blender code is necessary but far from sufficient for constructing structurally valid and physically coherent executable worlds.

\begin{table}[!htb]
\centering
\vspace{-6mm}
\caption{\textbf{Quantitative results on T1: Mechanism Executability.}}
\label{tab:t1}
\begingroup
\BTSetup{6}
\begin{tabular}{@{}>{\raggedright\arraybackslash}m{\BTmodelwidth}*{6}{>{\centering\arraybackslash}m{\BTbarwidth}}@{}}
\toprule
\multirow[c]{2}{\BTmodelwidth}[-\dimexpr(\aboverulesep+\belowrulesep+\cmidrulewidth)/2\relax]{\centering\textbf{Model}}
& \multicolumn{2}{c}{$Executability \uparrow$}
& \multicolumn{2}{c}{$Visual \uparrow$}
& \multicolumn{2}{c}{$Motion \uparrow$} \\
\cmidrule(lr){2-3}\cmidrule(lr){4-5}\cmidrule(lr){6-7}
& $E_1$ & $E_2$
& $V_1$ & $V_2$
& $M_1$ & $M_2$ \\
\midrule
\BTModel{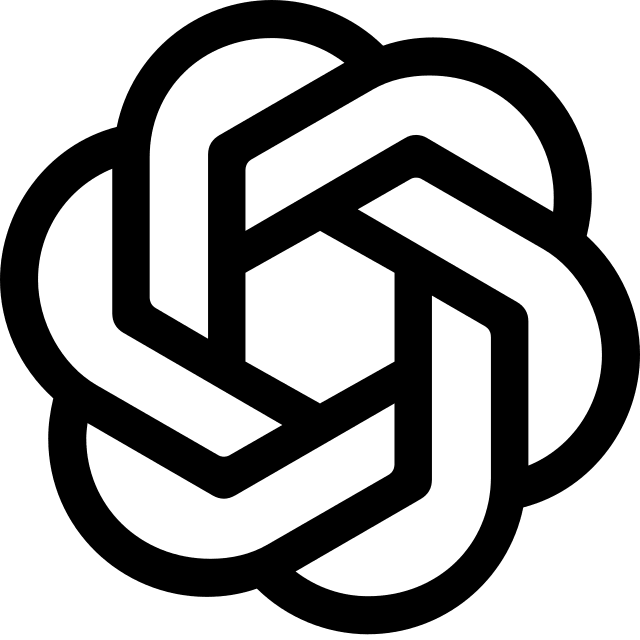}{GPT-6 Astra (Codex CLI)}{\cite{openai2026gpt6astra,openai_codex_cli}}
& \BTBar{BTExec}{100}{100}{\textbf{100\%}}
& \BTBar{BTExec}{100}{100}{\textbf{100\%}}
& \BTBar{BTVisual}{7.16}{10}{\textbf{7.16}}
& \BTBar{BTVisual}{7.16}{10}{\textbf{7.16}}
& \BTBar{BTMotion}{6.32}{10}{\textbf{6.32}}
& \BTBar{BTMotion}{6.32}{10}{\textbf{6.32}} \\
\BTModel{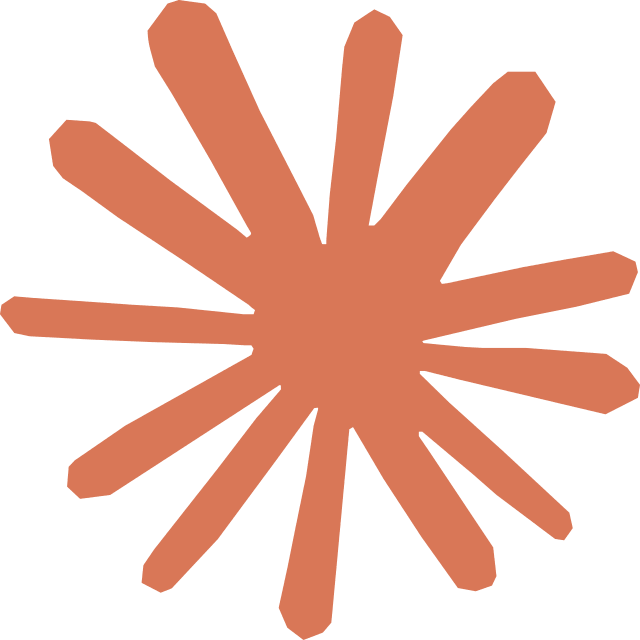}{Claude Fable 5.1 (Claude Code)}{\cite{anthropic2026fable51,anthropic_claude_code}}
& \BTBar{BTExec}{100}{100}{\textbf{100\%}}
& \BTBar{BTExec}{100}{100}{\textbf{100\%}}
& \BTBar{BTVisual}{6.44}{10}{6.44}
& \BTBar{BTVisual}{6.44}{10}{6.44}
& \BTBar{BTMotion}{5.68}{10}{5.68}
& \BTBar{BTMotion}{5.68}{10}{5.68} \\
\addlinespace[3pt]
\BTModel{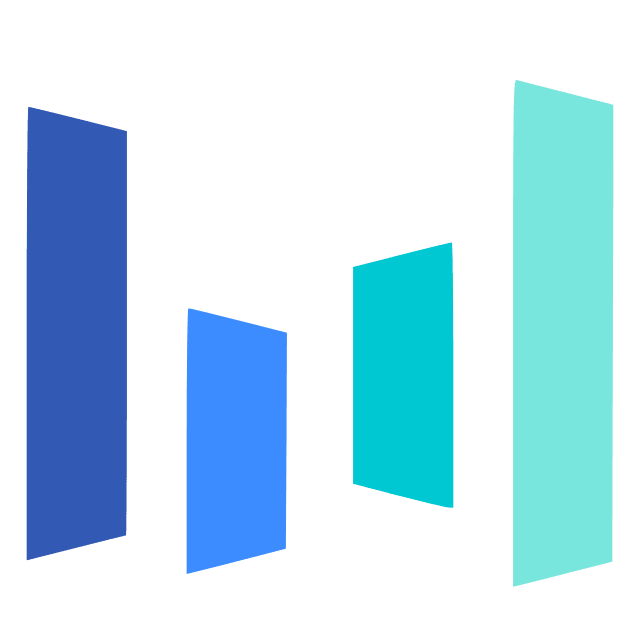}{Doubao-Seed-2.1-Pro}{\cite{bytedanceseed2026seed21}}
& \BTBar{BTExec}{8.00}{100}{8.00\%}
& \BTBar{BTExec}{82.00}{100}{82.00\%}
& \BTBar{BTVisual}{0.30}{10}{0.30}
& \BTBar{BTVisual}{2.16}{10}{2.16}
& \BTBar{BTMotion}{0.20}{10}{0.20}
& \BTBar{BTMotion}{1.04}{10}{1.04} \\
\BTModel{openai}{GPT-6 Astra}{\cite{openai2026gpt6astra}}
& \BTBar{BTExec}{58.00}{100}{58.00\%}
& \BTBar{BTExec}{94.00}{100}{94.00\%}
& \BTBar{BTVisual}{3.58}{10}{3.58}
& \BTBar{BTVisual}{5.84}{10}{5.84}
& \BTBar{BTMotion}{3.14}{10}{3.14}
& \BTBar{BTMotion}{5.08}{10}{5.08} \\
\BTModel{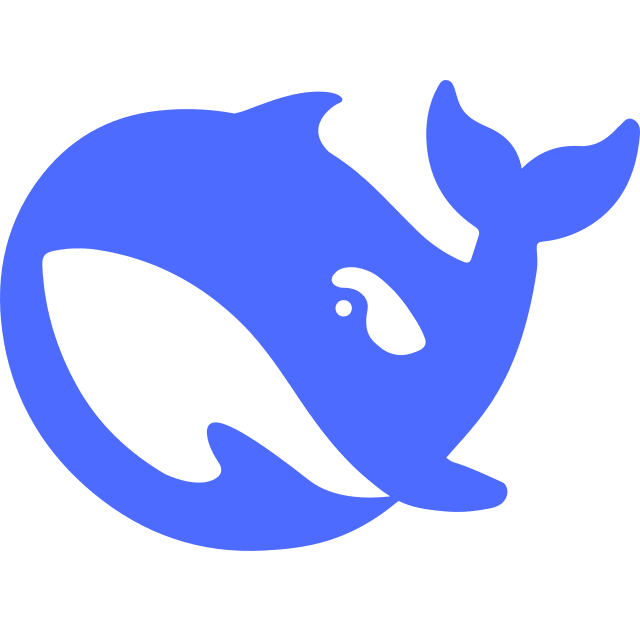}{DeepSeek-V4.1-Flash}{\cite{deepseekai2026deepseekv41flash}}
& \BTBar{BTExec}{14.00}{100}{14.00\%}
& \BTBar{BTExec}{82.00}{100}{82.00\%}
& \BTBar{BTVisual}{0.24}{10}{0.24}
& \BTBar{BTVisual}{1.76}{10}{1.76}
& \BTBar{BTMotion}{0.08}{10}{0.08}
& \BTBar{BTMotion}{0.78}{10}{0.78} \\
\BTModel{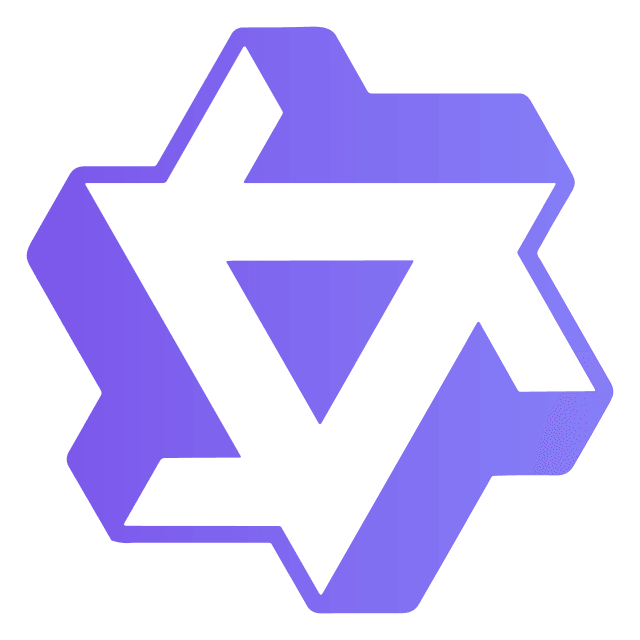}{Qwen3.8-Max}{\cite{alibaba2026qwen38max}}
& \BTBar{BTExec}{20.00}{100}{20.00\%}
& \BTBar{BTExec}{82.00}{100}{82.00\%}
& \BTBar{BTVisual}{0.34}{10}{0.34}
& \BTBar{BTVisual}{1.34}{10}{1.34}
& \BTBar{BTMotion}{0.18}{10}{0.18}
& \BTBar{BTMotion}{0.82}{10}{0.82} \\
\BTModel{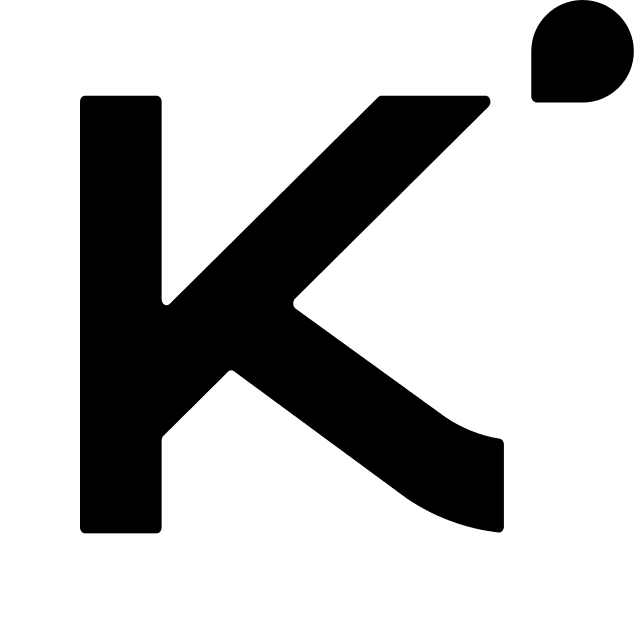}{Kimi K3}{\cite{kimiteam2026kimik3}}
& \BTBar{BTExec}{8.00}{100}{8.00\%}
& \BTBar{BTExec}{86.00}{100}{86.00\%}
& \BTBar{BTVisual}{0.12}{10}{0.12}
& \BTBar{BTVisual}{1.64}{10}{1.64}
& \BTBar{BTMotion}{0.04}{10}{0.04}
& \BTBar{BTMotion}{0.80}{10}{0.80} \\
\BTModel{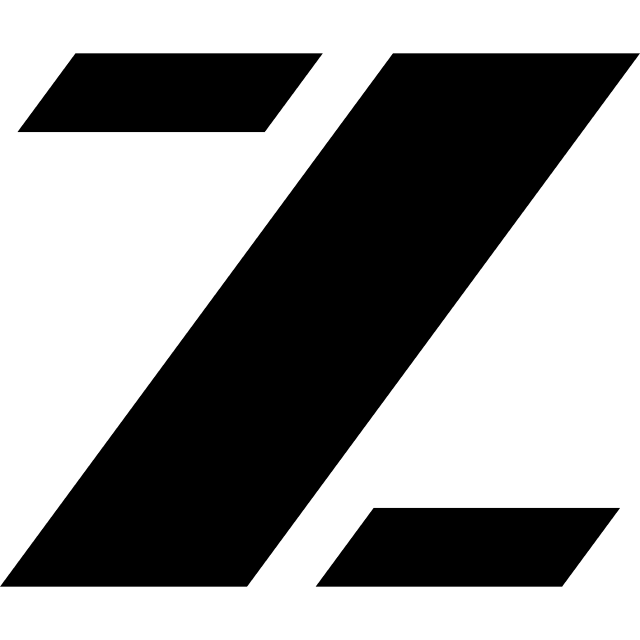}{GLM-5.3-Flash}{\cite{zai2026glm53flash}}
& \BTBar{BTExec}{26.00}{100}{26.00\%}
& \BTBar{BTExec}{78.00}{100}{78.00\%}
& \BTBar{BTVisual}{0.48}{10}{0.48}
& \BTBar{BTVisual}{1.52}{10}{1.52}
& \BTBar{BTMotion}{0.24}{10}{0.24}
& \BTBar{BTMotion}{0.84}{10}{0.84} \\
\BTModel{claude}{Claude Fable 5.1}{\cite{anthropic2026fable51}}
& \BTBar{BTExec}{18.00}{100}{18.00\%}
& \BTBar{BTExec}{100.00}{100}{\textbf{100.00\%}}
& \BTBar{BTVisual}{0.76}{10}{0.76}
& \BTBar{BTVisual}{4.04}{10}{4.04}
& \BTBar{BTMotion}{0.50}{10}{0.50}
& \BTBar{BTMotion}{3.24}{10}{3.24} \\
\bottomrule
\end{tabular}
\endgroup
\vspace{-4mm}
\end{table}

\vspace{-2mm}
\subsection{T1: Mechanism Executability.}
\vspace{-3mm}
Table~\ref{tab:t1} reports Visual Structure ($V$) and Mechanical Motion ($M$) on T1.
\textbf{Visual Structure and Mechanical Motion are distinct but asymmetric.}
Models with strong motion performance also achieve high structural scores, whereas strong structure does not always imply equally strong motion; for example, Fable 5.1 reaches $V=4.04$ but only $M=3.24$.
This suggests that strong motion execution depends on structural fidelity, yet structural correctness alone does not guarantee accurate mechanical dynamics.

\vspace{-6mm}
\subsection{T2: Compositional Executability}
\vspace{-4mm}
Table~\ref{tab:t2} reports Visual Structure ($V$) and Executable Causal Mechanism (ECM) on T2.
\textbf{Strong T1 motion performance does not necessarily transfer to multi-stage causal execution.}
For example, after repair, Seed-2.1 ranks fifth on T1 Mechanical Motion but seventh on T2 ECM.
This indicates that multi-stage execution requires capabilities beyond realizing individual mechanical motions.
% \vspace{-6mm}

\begin{table}
% [!htb]
[H]
\centering
\vspace{-5mm}
\caption{\textbf{Quantitative results on T2: Compositional Executability.}}
\label{tab:t2}
\begingroup
\BTSetup{6}
\begin{tabular}{@{}>{\raggedright\arraybackslash}m{\BTmodelwidth}*{6}{>{\centering\arraybackslash}m{\BTbarwidth}}@{}}
\toprule
\multirow[c]{2}{\BTmodelwidth}[-\dimexpr(\aboverulesep+\belowrulesep+\cmidrulewidth)/2\relax]{\centering\textbf{Model}}
& \multicolumn{2}{c}{$Executability \uparrow$}
& \multicolumn{2}{c}{$Visual \uparrow$}
& \multicolumn{2}{c}{$\mathrm{ECM} \uparrow$} \\
\cmidrule(lr){2-3}\cmidrule(lr){4-5}\cmidrule(lr){6-7}
& $E_1$ & $E_2$
& $V_1$ & $V_2$
& $\mathrm{ECM}_{1}$ & $\mathrm{ECM}_{2}$\\
\midrule
\BTModel{openai}{GPT-6 Astra (Codex CLI)}{\cite{openai2026gpt6astra,openai_codex_cli}}
& \BTBar{BTExec}{99}{100}{99\%}
& \BTBar{BTExec}{99}{100}{99\%}
& \BTBar{BTVisual}{5.33}{10}{\textbf{5.33}}
& \BTBar{BTVisual}{5.33}{10}{\textbf{5.33}}
& \BTBar{BTMotion}{5.25}{10}{\textbf{5.25}}
& \BTBar{BTMotion}{5.25}{10}{\textbf{5.25}} \\
\BTModel{claude}{Claude Fable 5.1 (Claude Code)}{\cite{anthropic2026fable51,anthropic_claude_code}}
& \BTBar{BTExec}{100}{100}{\textbf{100\%}}
& \BTBar{BTExec}{100}{100}{\textbf{100\%}}
& \BTBar{BTVisual}{4.45}{10}{4.45}
& \BTBar{BTVisual}{4.45}{10}{4.45}
& \BTBar{BTMotion}{4.08}{10}{4.08}
& \BTBar{BTMotion}{4.08}{10}{4.08} \\
\addlinespace[3pt]
\BTModel{bytedance}{Doubao-Seed-2.1-Pro}{\cite{bytedanceseed2026seed21}}
& \BTBar{BTExec}{2.00}{100}{2.00\%}
& \BTBar{BTExec}{73.00}{100}{73.00\%}
& \BTBar{BTVisual}{0.04}{10}{0.04}
& \BTBar{BTVisual}{1.12}{10}{1.12}
& \BTBar{BTMotion}{0.00}{10}{0.00}
& \BTBar{BTMotion}{0.01}{10}{0.01} \\
\BTModel{openai}{GPT-6 Astra}{\cite{openai2026gpt6astra}}
& \BTBar{BTExec}{81.00}{100}{81.00\%}
& \BTBar{BTExec}{91.00}{100}{91.00\%}
& \BTBar{BTVisual}{3.46}{10}{3.46}
& \BTBar{BTVisual}{3.86}{10}{3.86}
& \BTBar{BTMotion}{2.23}{10}{2.23}
& \BTBar{BTMotion}{2.47}{10}{2.47} \\
\BTModel{deepseek}{DeepSeek-V4.1-Flash}{\cite{deepseekai2026deepseekv41flash}}
& \BTBar{BTExec}{2.00}{100}{2.00\%}
& \BTBar{BTExec}{88.00}{100}{88.00\%}
& \BTBar{BTVisual}{0.05}{10}{0.05}
& \BTBar{BTVisual}{1.83}{10}{1.83}
& \BTBar{BTMotion}{0.00}{10}{0.00}
& \BTBar{BTMotion}{0.09}{10}{0.09} \\
\BTModel{qwen}{Qwen3.8-Max}{\cite{alibaba2026qwen38max}}
& \BTBar{BTExec}{9.00}{100}{9.00\%}
& \BTBar{BTExec}{62.00}{100}{62.00\%}
& \BTBar{BTVisual}{0.06}{10}{0.06}
& \BTBar{BTVisual}{1.21}{10}{1.21}
& \BTBar{BTMotion}{0.00}{10}{0.00}
& \BTBar{BTMotion}{0.00}{10}{0.00} \\
\BTModel{kimi}{Kimi K3}{\cite{kimiteam2026kimik3}}
& \BTBar{BTExec}{10.00}{100}{10.00\%}
& \BTBar{BTExec}{79.00}{100}{79.00\%}
& \BTBar{BTVisual}{0.19}{10}{0.19}
& \BTBar{BTVisual}{1.62}{10}{1.62}
& \BTBar{BTMotion}{0.00}{10}{0.00}
& \BTBar{BTMotion}{0.00}{10}{0.00} \\
\BTModel{zai}{GLM-5.3-Flash}{\cite{zai2026glm53flash}}
& \BTBar{BTExec}{22.00}{100}{22.00\%}
& \BTBar{BTExec}{79.00}{100}{79.00\%}
& \BTBar{BTVisual}{0.55}{10}{0.55}
& \BTBar{BTVisual}{1.97}{10}{1.97}
& \BTBar{BTMotion}{0.07}{10}{0.07}
& \BTBar{BTMotion}{0.15}{10}{0.15} \\
\BTModel{claude}{Claude Fable 5.1}{\cite{anthropic2026fable51}}
& \BTBar{BTExec}{91.00}{100}{91.00\%}
& \BTBar{BTExec}{93.00}{100}{93.00\%}
& \BTBar{BTVisual}{3.42}{10}{3.42}
& \BTBar{BTVisual}{3.49}{10}{3.49}
& \BTBar{BTMotion}{1.83}{10}{1.83}
& \BTBar{BTMotion}{1.84}{10}{1.84} \\
\bottomrule
\end{tabular}
\endgroup
\vspace{-8mm}
\end{table}

\vspace{-1mm}
\subsection{T3: Intent-Driven Executable World Synthesis}
\vspace{-3mm}
Table~\ref{tab:t3-1} reports Visual Structure ($V$) and Executable Causal Mechanism (ECM) for T3-1.
Table~\ref{tab:t3-2} reports Visual Structure ($V$), Executable Causal Mechanism (ECM) and Goal Satisfaction (\textbf{G}) for T3-2.

\noindent\textbf{{Component-Constrained Synthesis.}}
\textbf{This task remains challenging even when the required components are explicitly provided.}
After constrained repair, only the harness-based models achieve relatively strong ECM scores.
Among API-based models, even GPT-6 Astra and Fable 5.1 reach only $3.34$ and $2.75$, respectively, while all remaining models score at most $0.40$.
These results show that, even with the full component set provided, models still struggle to organize the components into a spatially coherent and physically valid causal system. 

\vspace{-0.5mm}
\noindent\textbf{{Goal-Constrained Synthesis.}}
\textbf{Removing explicit component-level guidance further increases the difficulty of executable world synthesis.}
Compared with Component-Constrained Synthesis, models achieve substantially lower ECM performance when only high-level intent is provided.
This degradation suggests that inferring a suitable mechanism introduces substantial additional difficulty beyond spatially organizing and causally connecting predefined components.

\begin{table}
[!htb]
% [H]
\centering
\vspace{-6.0mm}
\caption{\textbf{Quantitative results on T3-1: Component-Constrained Synthesis.}}
\label{tab:t3-1}
\begingroup
\BTSetup{6}
\begin{tabular}{@{}>{\raggedright\arraybackslash}m{\BTmodelwidth}*{6}{>{\centering\arraybackslash}m{\BTbarwidth}}@{}}
\toprule
\multirow[c]{2}{\BTmodelwidth}[-\dimexpr(\aboverulesep+\belowrulesep+\cmidrulewidth)/2\relax]{\centering\textbf{Model}}
& \multicolumn{2}{c}{$Executability \uparrow$}
& \multicolumn{2}{c}{$Visual \uparrow$}
& \multicolumn{2}{c}{$\mathrm{ECM} \uparrow$} \\
\cmidrule(lr){2-3}\cmidrule(lr){4-5}\cmidrule(lr){6-7}
& $E_1$ & $E_2$
& $V_1$ & $V_2$
& $\mathrm{ECM}_{1}$ & $\mathrm{ECM}_{2}$\\
\midrule
\BTModel{openai}{GPT-6 Astra (Codex CLI)}{\cite{openai2026gpt6astra,openai_codex_cli}}
& \BTBar{BTExec}{100}{100}{\textbf{100\%}}
& \BTBar{BTExec}{100}{100}{\textbf{100\%}}
& \BTBar{BTVisual}{5.67}{10}{\textbf{5.67}}
& \BTBar{BTVisual}{5.67}{10}{\textbf{5.67}}
& \BTBar{BTMotion}{6.11}{10}{\textbf{6.11}}
& \BTBar{BTMotion}{6.11}{10}{\textbf{6.11}} \\
\BTModel{claude}{Claude Fable 5.1 (Claude Code)}{\cite{anthropic2026fable51,anthropic_claude_code}}
& \BTBar{BTExec}{100}{100}{\textbf{100\%}}
& \BTBar{BTExec}{100}{100}{\textbf{100\%}}
& \BTBar{BTVisual}{5.21}{10}{5.21}
& \BTBar{BTVisual}{5.21}{10}{5.21}
& \BTBar{BTMotion}{5.70}{10}{5.70}
& \BTBar{BTMotion}{5.70}{10}{5.70} \\
\addlinespace[3pt]
\BTModel{bytedance}{Doubao-Seed-2.1-Pro}{\cite{bytedanceseed2026seed21}}
& \BTBar{BTExec}{9.52}{100}{9.52\%}
& \BTBar{BTExec}{84.13}{100}{84.13\%}
& \BTBar{BTVisual}{0.33}{10}{0.33}
& \BTBar{BTVisual}{2.24}{10}{2.24}
& \BTBar{BTMotion}{0.05}{10}{0.05}
& \BTBar{BTMotion}{0.13}{10}{0.13} \\
\BTModel{openai}{GPT-6 Astra}{\cite{openai2026gpt6astra}}
& \BTBar{BTExec}{44.44}{100}{44.44\%}
& \BTBar{BTExec}{85.71}{100}{85.71\%}
& \BTBar{BTVisual}{2.22}{10}{2.22}
& \BTBar{BTVisual}{4.08}{10}{4.08}
& \BTBar{BTMotion}{1.83}{10}{1.83}
& \BTBar{BTMotion}{3.34}{10}{3.34} \\
\BTModel{deepseek}{DeepSeek-V4.1-Flash}{\cite{deepseekai2026deepseekv41flash}}
& \BTBar{BTExec}{4.76}{100}{4.76\%}
& \BTBar{BTExec}{87.30}{100}{87.30\%}
& \BTBar{BTVisual}{0.10}{10}{0.10}
& \BTBar{BTVisual}{2.11}{10}{2.11}
& \BTBar{BTMotion}{0.07}{10}{0.07}
& \BTBar{BTMotion}{0.29}{10}{0.29} \\
\BTModel{qwen}{Qwen3.8-Max}{\cite{alibaba2026qwen38max}}
& \BTBar{BTExec}{0.00}{100}{0.00\%}
& \BTBar{BTExec}{84.13}{100}{84.13\%}
& \BTBar{BTVisual}{0.00}{10}{0.00}
& \BTBar{BTVisual}{1.62}{10}{1.62}
& \BTBar{BTMotion}{0.00}{10}{0.00}
& \BTBar{BTMotion}{0.03}{10}{0.03} \\
\BTModel{kimi}{Kimi K3}{\cite{kimiteam2026kimik3}}
& \BTBar{BTExec}{6.35}{100}{6.35\%}
& \BTBar{BTExec}{95.24}{100}{95.24\%}
& \BTBar{BTVisual}{0.17}{10}{0.17}
& \BTBar{BTVisual}{2.37}{10}{2.37}
& \BTBar{BTMotion}{0.01}{10}{0.01}
& \BTBar{BTMotion}{0.40}{10}{0.40} \\
\BTModel{zai}{GLM-5.3-Flash}{\cite{zai2026glm53flash}}
& \BTBar{BTExec}{20.63}{100}{20.63\%}
& \BTBar{BTExec}{71.43}{100}{71.43\%}
& \BTBar{BTVisual}{0.67}{10}{0.67}
& \BTBar{BTVisual}{2.33}{10}{2.33}
& \BTBar{BTMotion}{0.10}{10}{0.10}
& \BTBar{BTMotion}{0.39}{10}{0.39} \\
\BTModel{claude}{Claude Fable 5.1}{\cite{anthropic2026fable51}}
& \BTBar{BTExec}{98.41}{100}{98.41\%}
& \BTBar{BTExec}{100.00}{100}{\textbf{100.00\%}}
& \BTBar{BTVisual}{4.32}{10}{4.32}
& \BTBar{BTVisual}{4.38}{10}{4.38}
& \BTBar{BTMotion}{2.73}{10}{2.73}
& \BTBar{BTMotion}{2.75}{10}{2.75} \\
\bottomrule
\end{tabular}
\endgroup
\vspace{-4mm}
\end{table}

\textbf{Models satisfy initial-state requirements much more reliably than behavioral and final-state requirements.}
By analyzing the satisfaction rates of the initial state (I), intermediate behavior (B), and final state (F), we find that across API-based models, initial-state requirements are generally satisfied more reliably than behavioral and final-state requirements.
This suggests that models can often instantiate a plausible initial configuration, but struggle to execute the required causal process and carry it through to the desired final state.
The gap highlights a core challenge of executable world modeling: constructing a valid initial world is insufficient unless the underlying mechanism can sustain correct causal interactions over time and reliably produce the intended outcome.
Additional visualizations are provided in Appendix.~\ref{sec:app1}.

\begin{table}
% [!htb]
[H]
\centering
\vspace{-3mm}
\caption{\textbf{Quantitative results on T3-2: Goal-Constrained Synthesis.}}
\label{tab:t3-2}
\begingroup
\BTSetup{8}
\begin{tabular}{@{}>{\raggedright\arraybackslash}m{\BTmodelwidth}*{8}{>{\centering\arraybackslash}m{\BTbarwidth}}@{}}
\toprule
\multirow[c]{2}{\BTmodelwidth}[-\dimexpr(\aboverulesep+\belowrulesep+\cmidrulewidth)/2\relax]{\centering\textbf{Model}}
& \multicolumn{2}{c}{$Executability \uparrow$}
& \multicolumn{2}{c}{$Visual \uparrow$}
& \multicolumn{2}{c}{$\mathrm{ECM} \uparrow$}
& \multicolumn{2}{c}{Goal Satisfaction $\uparrow$} \\
\cmidrule(lr){2-3}\cmidrule(lr){4-5}\cmidrule(lr){6-7}\cmidrule(lr){8-9}
& $E_1$ & $E_2$
& $V_1$ & $V_2$
& $\mathrm{ECM}_{1}$ & $\mathrm{ECM}_{2}$
& $\mathbf{G}_{1}$ & $\mathbf{G}_{2}$ \\
\midrule
\BTModel{openai}{GPT-6 Astra (Codex CLI)}{\cite{openai2026gpt6astra,openai_codex_cli}}
& \BTBar{BTExec}{100}{100}{\textbf{100\%}}
& \BTBar{BTExec}{100}{100}{\textbf{100\%}}
& \BTBar{BTVisual}{6.28}{10}{\textbf{6.28}}
& \BTBar{BTVisual}{6.28}{10}{\textbf{6.28}}
& \BTBar{BTMotion}{5.82}{10}{\textbf{5.82}}
& \BTBar{BTMotion}{5.82}{10}{\textbf{5.82}}
& \BTBar{BTGoal}{90.14}{100}{\textbf{90.14\%}}
& \BTBar{BTGoal}{90.14}{100}{\textbf{90.14\%}} \\
\BTModel{claude}{Claude Fable 5.1 (Claude Code)}{\cite{anthropic2026fable51,anthropic_claude_code}}
& \BTBar{BTExec}{100}{100}{\textbf{100\%}}
& \BTBar{BTExec}{100}{100}{\textbf{100\%}}
& \BTBar{BTVisual}{5.83}{10}{5.83}
& \BTBar{BTVisual}{5.83}{10}{5.83}
& \BTBar{BTMotion}{5.17}{10}{5.17}
& \BTBar{BTMotion}{5.17}{10}{5.17}
& \BTBar{BTGoal}{75.19}{100}{75.19\%}
& \BTBar{BTGoal}{75.19}{100}{75.19\%} \\
\addlinespace[3pt]
\BTModel{bytedance}{Doubao-Seed-2.1-Pro}{\cite{bytedanceseed2026seed21}}
& \BTBar{BTExec}{1.67}{100}{1.67\%}
& \BTBar{BTExec}{70.00}{100}{70.00\%}
& \BTBar{BTVisual}{0.05}{10}{0.05}
& \BTBar{BTVisual}{1.82}{10}{1.82}
& \BTBar{BTMotion}{0.02}{10}{0.02}
& \BTBar{BTMotion}{0.33}{10}{0.33}
& \BTBar{BTGoal}{0.26}{100}{0.26\%}
& \BTBar{BTGoal}{7.48}{100}{7.48\%} \\
\BTModel{openai}{GPT-6 Astra}{\cite{openai2026gpt6astra}}
& \BTBar{BTExec}{40.00}{100}{40.00\%}
& \BTBar{BTExec}{91.67}{100}{91.67\%}
& \BTBar{BTVisual}{1.97}{10}{1.97}
& \BTBar{BTVisual}{4.33}{10}{4.33}
& \BTBar{BTMotion}{1.15}{10}{1.15}
& \BTBar{BTMotion}{2.55}{10}{2.55}
& \BTBar{BTGoal}{19.17}{100}{19.17\%}
& \BTBar{BTGoal}{41.07}{100}{41.07\%} \\
\BTModel{deepseek}{DeepSeek-V4.1-Flash}{\cite{deepseekai2026deepseekv41flash}}
& \BTBar{BTExec}{10.00}{100}{10.00\%}
& \BTBar{BTExec}{81.67}{100}{81.67\%}
& \BTBar{BTVisual}{0.25}{10}{0.25}
& \BTBar{BTVisual}{1.77}{10}{1.77}
& \BTBar{BTMotion}{0.05}{10}{0.05}
& \BTBar{BTMotion}{0.52}{10}{0.52}
& \BTBar{BTGoal}{1.15}{100}{1.15\%}
& \BTBar{BTGoal}{7.52}{100}{7.52\%} \\
\BTModel{qwen}{Qwen3.8-Max}{\cite{alibaba2026qwen38max}}
& \BTBar{BTExec}{0.00}{100}{0.00\%}
& \BTBar{BTExec}{81.67}{100}{81.67\%}
& \BTBar{BTVisual}{0.00}{10}{0.00}
& \BTBar{BTVisual}{1.83}{10}{1.83}
& \BTBar{BTMotion}{0.00}{10}{0.00}
& \BTBar{BTMotion}{0.37}{10}{0.37}
& \BTBar{BTGoal}{0.00}{100}{0.00\%}
& \BTBar{BTGoal}{5.85}{100}{5.85\%} \\
\BTModel{kimi}{Kimi K3}{\cite{kimiteam2026kimik3}}
& \BTBar{BTExec}{3.33}{100}{3.33\%}
& \BTBar{BTExec}{81.67}{100}{81.67\%}
& \BTBar{BTVisual}{0.13}{10}{0.13}
& \BTBar{BTVisual}{2.70}{10}{2.70}
& \BTBar{BTMotion}{0.05}{10}{0.05}
& \BTBar{BTMotion}{0.63}{10}{0.63}
& \BTBar{BTGoal}{1.53}{100}{1.53\%}
& \BTBar{BTGoal}{18.13}{100}{18.13\%} \\
\BTModel{zai}{GLM-5.3-Flash}{\cite{zai2026glm53flash}}
& \BTBar{BTExec}{10.00}{100}{10.00\%}
& \BTBar{BTExec}{30.00}{100}{30.00\%}
& \BTBar{BTVisual}{0.32}{10}{0.32}
& \BTBar{BTVisual}{1.07}{10}{1.07}
& \BTBar{BTMotion}{0.00}{10}{0.00}
& \BTBar{BTMotion}{0.15}{10}{0.15}
& \BTBar{BTGoal}{0.86}{100}{0.86\%}
& \BTBar{BTGoal}{5.29}{100}{5.29\%} \\
\BTModel{claude}{Claude Fable 5.1}{\cite{anthropic2026fable51}}
& \BTBar{BTExec}{73.33}{100}{73.33\%}
& \BTBar{BTExec}{78.33}{100}{78.33\%}
& \BTBar{BTVisual}{3.22}{10}{3.22}
& \BTBar{BTVisual}{3.47}{10}{3.47}
& \BTBar{BTMotion}{1.68}{10}{1.68}
& \BTBar{BTMotion}{1.85}{10}{1.85}
& \BTBar{BTGoal}{28.83}{100}{28.83\%}
& \BTBar{BTGoal}{31.31}{100}{31.31\%} \\
\bottomrule
\end{tabular}
\endgroup
\vspace{-5mm}
\end{table}

\vspace{-3mm}
\subsection{Long-horizon Analysis}
\begin{figure}[!htb]
% [H]
    \centering
    \includegraphics[width=1.0\textwidth]{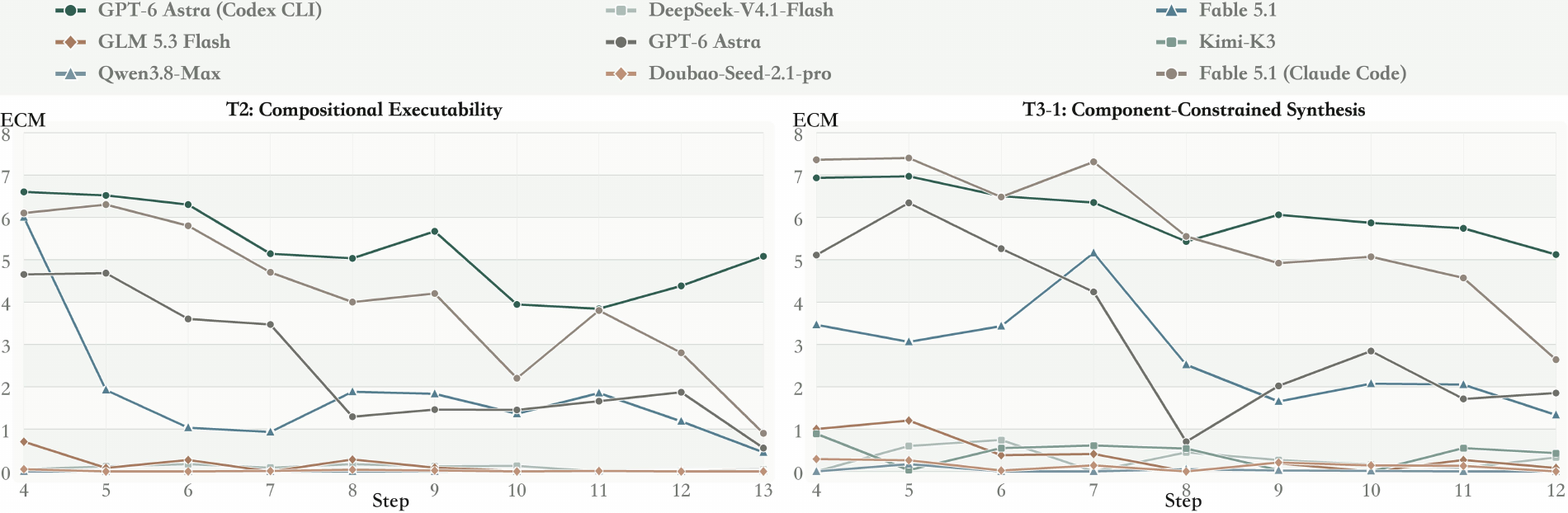}
    \vspace{-4mm}
\caption{\textbf{Long-horizon execution performance across stages.}}
    % \vspace{-2mm}
    \label{fig:step}
    \vspace{-4mm}
\end{figure}

\vspace{-3mm}
We further analyze long-horizon execution by grouping the repaired results according to the number of execution steps, which is shown in Fig.~\ref{fig:step}.

\textbf{Performance exhibits a decreasing trend as the execution horizon increases.}
On T2, the average ECM across models decreases steadily with the number of execution stages.
Fable 5.1 exhibits a sharp drop around 5 steps, while Fable 5.1 (Claude Code) and GPT-6 Astra decline more gradually, with their 13-step performance falling to roughly one-fifth of their short-horizon scores.
GPT-6 Astra (Codex CLI) shows the smallest degradation and remains comparatively stable across longer sequences. 

A similar trend appears on T3-1.
The four models discussed above exhibit declining performance as the number of stages increases, with GPT-6 Astra (Codex CLI) again exhibiting the smallest drop.
GPT-6 Astra shows a particularly pronounced decline between 5 and 8 steps, falling below $1.0$ ECM at 8 steps.
These results indicate that maintaining mechanically valid causal propagation becomes increasingly difficult as more stages must be connected, highlighting long-horizon execution as a significant challenge for current executable-world modeling systems.

\vspace{-2mm}

\begin{figure}[!htb]
% [!htbp]
% [H]
    \centering
    \includegraphics[width=1.0\linewidth]{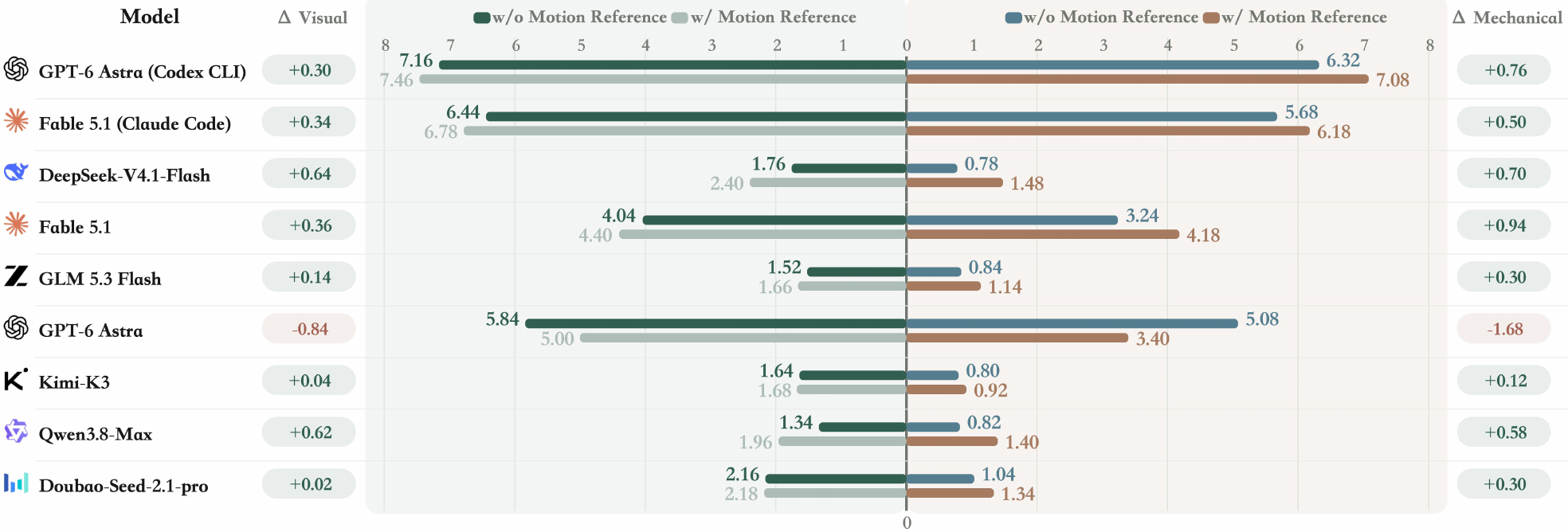}
\caption{\textbf{Effect of motion references on T1.}}
    \label{fig:motion_ref_t1}
    \vspace{-2mm}
\end{figure}

\begin{figure}[!htb]
% [!htbp]
% [H]
    \centering
    \includegraphics[width=1.0\linewidth]{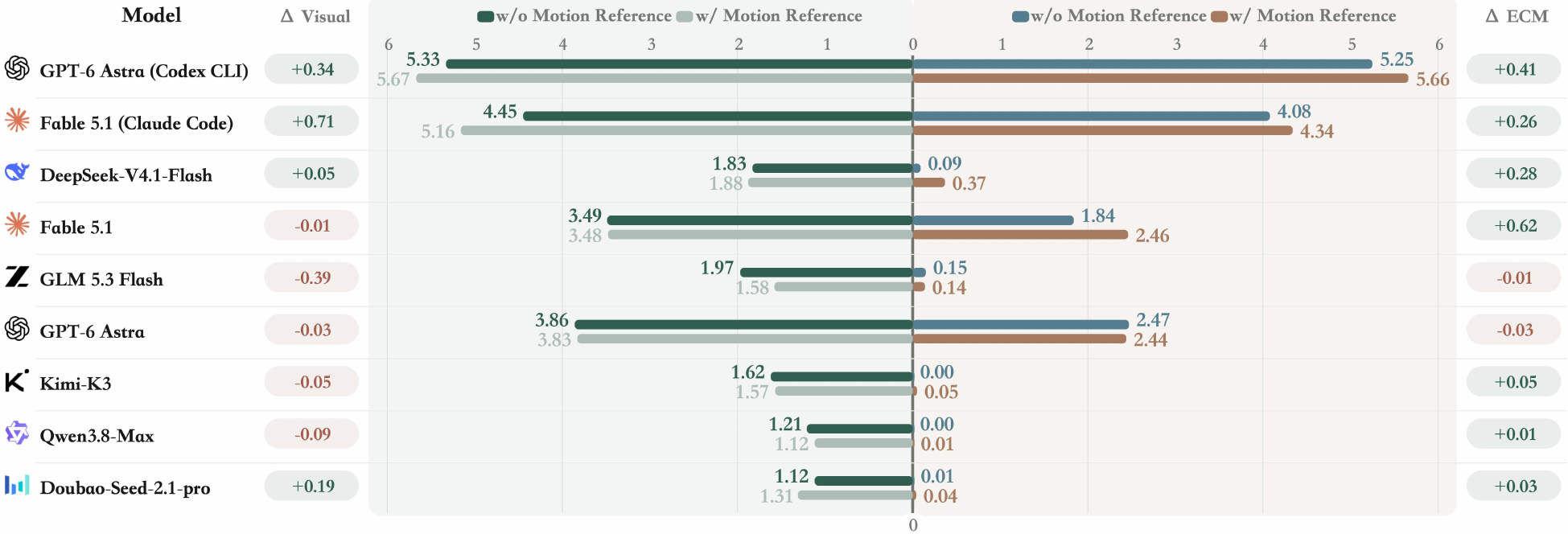}
\caption{\textbf{Effect of motion references on T2.}}
    \label{fig:motion_ref_t2}
    \vspace{-5mm}
\end{figure}

% \vspace{-3mm}
\subsection{Effect of Motion References}
% \vspace{-3mm}
We further investigate whether augmenting static visual input with frames extracted from the corresponding reference video improves model performance.
On T1, motion references improve performance for all models except GPT-6 Astra, although the gains are generally modest, as shown in Fig.~\ref{fig:motion_ref_t1}.
On T2, most models also benefit from additional motion references, while GLM-5.3-Flash and GPT-6 Astra exhibit marginal decreases, as shown in Fig.~\ref{fig:motion_ref_t2}.

Overall, the improvements remain limited.
This suggests that effectively interpreting motion evidence and translating it into executable mechanical structures remains challenging for current models.

\vspace{-3mm}
\section{Conclusion}
\vspace{-3mm}
We introduce \textbf{ExeWorldBench}, a benchmark for evaluating executable 3D world modeling with code across mechanism execution, multi-stage causal composition, and intent-driven synthesis.
Our results show that high code executability does not necessarily translate into strong structural, mechanical, or causal correctness, and that performance generally declines as causal complexity and execution horizon increase. 
We hope ExeWorldBench encourages progress toward models that can construct physically coherent executable worlds.

\vspace{0.25cm}

% \clearpage
\subsection*{AI use statement}
In this work, we used generative AI tools for (i) benchmark data construction and evaluation, and (ii) language polishing of the manuscript. 
We have not used generative AI tools for research ideation, core methodological design, or drawing scientific conclusions. 
The AI-assisted outputs were reviewed by the authors, who take full responsibility for the final content of the paper.

\bibliography{iclr2027_conference}
\bibliographystyle{iclr2027_conference}

\clearpage

\appendix
\section{Appendix}

\subsection{Data Construction Details}
Tab.~\ref{tab:component_usage} presents component usage statistics for T2.

\begin{table}[!tbp]
\centering
\caption{Component usage statistics.}
\label{tab:component_usage}
\begin{tabular}{lrr}
\toprule
\textbf{Component Category} & \textbf{Count} & \textbf{Percentage} \\
\midrule
Ramp                      & 256 & 30.12\% \\
Catch Bowl                         & 100 & 11.76\% \\
Domino Chain                  & 79  & 9.29\%  \\
Funnel                & 60  & 7.06\%  \\
Seesaw                      & 54  & 6.35\%  \\
L-Shaped Chute                     & 53  & 6.24\%  \\
Three-Segment Switchback Track                   & 51  & 6.00\%  \\
Four-Paddle Wheel                  & 46  & 5.41\%  \\
Two-Rod Revolute Linkage           & 46  & 5.41\%  \\
Deflector Plate       & 41  & 4.82\%  \\
Pendulum                           & 28  & 3.29\%  \\
Pegboard                           & 17  & 2.00\%  \\
Single-Pulley System                     & 12  & 1.41\%  \\
Horizontal Board  & 4   & 0.47\%  \\
Multi-Ball Staircase               & 3   & 0.35\%  \\
\midrule
\textbf{Total}                     & \textbf{850} & \textbf{100\%} \\
\bottomrule
\end{tabular}
\end{table}

% \subsection{Metric}

\subsection{More Result}
\label{sec:app1}

\begin{figure*}[t]
% \vspace{-10mm}
\includegraphics[width=0.5\linewidth]{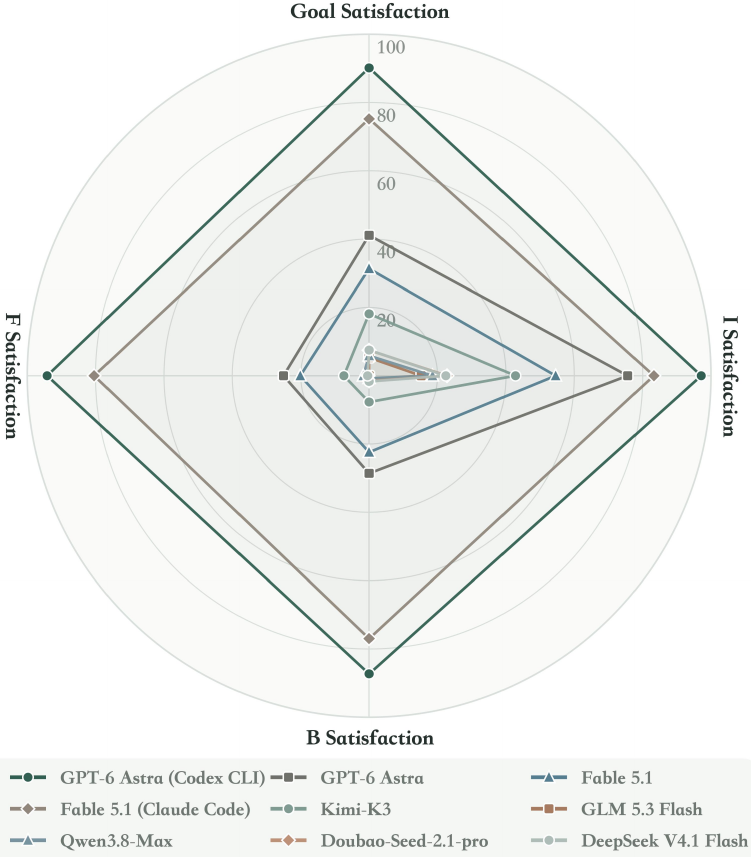}
  \centering
  % \vspace{-0.9em}  
\caption{\textbf{Goal satisfaction on T3-2.}}
% \vspace{-1.4em}
\label{fig:goal}
\end{figure*}

By analyzing the satisfaction rates of the initial state (I), intermediate behavior (B), and final state (F), we illustrate the detailed result of Goal Satisfaction rate in Fig.~\ref{fig:goal}.

We additionally report conditioned results of API-based models, computed only over code-executable cases, for all tasks in Tab.~\ref{tab:app_t1}, Tab.~\ref{tab:app_t2}, Tab.~\ref{tab:app_t3-1}, and Tab.~\ref{tab:app_t3-2}.

\begin{table}[!htbp]
\centering
\caption{\textbf{Quantitative results on T1: Mechanism Executability.}}
\label{tab:app_t1}
\begingroup
\BTSetup{6}
\begin{tabular}{@{}>{\raggedright\arraybackslash}m{\BTmodelwidth}*{6}{>{\centering\arraybackslash}m{\BTbarwidth}}@{}}
\toprule
\multirow[c]{2}{\BTmodelwidth}[-\dimexpr(\aboverulesep+\belowrulesep+\cmidrulewidth)/2\relax]{\centering\textbf{Model}}
& \multicolumn{2}{c}{$Executability \uparrow$}
& \multicolumn{2}{c}{$Visual \uparrow$}
& \multicolumn{2}{c}{$Motion \uparrow$} \\
\cmidrule(lr){2-3}\cmidrule(lr){4-5}\cmidrule(lr){6-7}
& $E_1$ & $E_2$
& $V_1$ & $V_2$
& $M_1$ & $M_2$ \\
\midrule
\BTModel{bytedance}{Doubao-Seed-2.1-Pro}{\cite{bytedanceseed2026seed21}}
& \BTBar{BTExec}{8.00}{100}{8.00\%}
& \BTBar{BTExec}{82.00}{100}{82.00\%}
& \BTBar{BTVisual}{3.75}{10}{3.75}
& \BTBar{BTVisual}{2.63}{10}{2.63}
& \BTBar{BTMotion}{2.50}{10}{2.50}
& \BTBar{BTMotion}{1.27}{10}{1.27} \\
\BTModel{openai}{GPT-6 Astra}{\cite{openai2026gpt6astra}}
& \BTBar{BTExec}{58.00}{100}{\textbf{58.00\%}}
& \BTBar{BTExec}{94.00}{100}{94.00\%}
& \BTBar{BTVisual}{6.17}{10}{\textbf{6.17}}
& \BTBar{BTVisual}{6.21}{10}{\textbf{6.21}}
& \BTBar{BTMotion}{5.41}{10}{\textbf{5.41}}
& \BTBar{BTMotion}{5.40}{10}{\textbf{5.40}} \\
\BTModel{deepseek}{DeepSeek-V4.1-Flash}{\cite{deepseekai2026deepseekv41flash}}
& \BTBar{BTExec}{14.00}{100}{14.00\%}
& \BTBar{BTExec}{82.00}{100}{82.00\%}
& \BTBar{BTVisual}{1.71}{10}{1.71}
& \BTBar{BTVisual}{2.15}{10}{2.15}
& \BTBar{BTMotion}{0.57}{10}{0.57}
& \BTBar{BTMotion}{0.95}{10}{0.95} \\
\BTModel{qwen}{Qwen3.8-Max}{\cite{alibaba2026qwen38max}}
& \BTBar{BTExec}{20.00}{100}{20.00\%}
& \BTBar{BTExec}{82.00}{100}{82.00\%}
& \BTBar{BTVisual}{1.70}{10}{1.70}
& \BTBar{BTVisual}{1.63}{10}{1.63}
& \BTBar{BTMotion}{0.90}{10}{0.90}
& \BTBar{BTMotion}{1.00}{10}{1.00} \\
\BTModel{kimi}{Kimi K3}{\cite{kimiteam2026kimik3}}
& \BTBar{BTExec}{8.00}{100}{8.00\%}
& \BTBar{BTExec}{86.00}{100}{86.00\%}
& \BTBar{BTVisual}{1.50}{10}{1.50}
& \BTBar{BTVisual}{1.91}{10}{1.91}
& \BTBar{BTMotion}{0.50}{10}{0.50}
& \BTBar{BTMotion}{0.93}{10}{0.93} \\
\BTModel{zai}{GLM-5.3-Flash}{\cite{zai2026glm53flash}}
& \BTBar{BTExec}{26.00}{100}{26.00\%}
& \BTBar{BTExec}{78.00}{100}{78.00\%}
& \BTBar{BTVisual}{1.85}{10}{1.85}
& \BTBar{BTVisual}{1.95}{10}{1.95}
& \BTBar{BTMotion}{0.92}{10}{0.92}
& \BTBar{BTMotion}{1.08}{10}{1.08} \\
\BTModel{claude}{Claude Fable 5.1}{\cite{anthropic2026fable51}}
& \BTBar{BTExec}{18.00}{100}{18.00\%}
& \BTBar{BTExec}{100.00}{100}{\textbf{100.00\%}}
& \BTBar{BTVisual}{4.22}{10}{4.22}
& \BTBar{BTVisual}{4.04}{10}{4.04}
& \BTBar{BTMotion}{2.78}{10}{2.78}
& \BTBar{BTMotion}{3.24}{10}{3.24} \\
\bottomrule
\end{tabular}
\endgroup
\end{table}

\begin{table}[!htbp]
\centering
\caption{\textbf{Quantitative results on T2: Compositional Executability.}}
\label{tab:app_t2}
\begingroup
\BTSetup{6}
\begin{tabular}{@{}>{\raggedright\arraybackslash}m{\BTmodelwidth}*{6}{>{\centering\arraybackslash}m{\BTbarwidth}}@{}}
\toprule
\multirow[c]{2}{\BTmodelwidth}[-\dimexpr(\aboverulesep+\belowrulesep+\cmidrulewidth)/2\relax]{\centering\textbf{Model}}
& \multicolumn{2}{c}{$Executability \uparrow$}
& \multicolumn{2}{c}{$Visual \uparrow$}
& \multicolumn{2}{c}{$\mathrm{ECM} \uparrow$} \\
\cmidrule(lr){2-3}\cmidrule(lr){4-5}\cmidrule(lr){6-7}
& $E_1$ & $E_2$
& $V_1$ & $V_2$
& $\mathrm{ECM}_{1}$ & $\mathrm{ECM}_{2}$ \\
\midrule
\BTModel{bytedance}{Doubao-Seed-2.1-Pro}{\cite{bytedanceseed2026seed21}}
& \BTBar{BTExec}{2.00}{100}{2.00\%}
& \BTBar{BTExec}{73.00}{100}{73.00\%}
& \BTBar{BTVisual}{2.00}{10}{2.00}
& \BTBar{BTVisual}{1.53}{10}{1.53}
& \BTBar{BTMotion}{0.00}{10}{0.00}
& \BTBar{BTMotion}{0.02}{10}{0.02} \\
\BTModel{openai}{GPT-6 Astra}{\cite{openai2026gpt6astra}}
& \BTBar{BTExec}{81.00}{100}{81.00\%}
& \BTBar{BTExec}{91.00}{100}{91.00\%}
& \BTBar{BTVisual}{4.27}{10}{\textbf{4.27}}
& \BTBar{BTVisual}{4.24}{10}{\textbf{4.24}}
& \BTBar{BTMotion}{2.75}{10}{\textbf{2.75}}
& \BTBar{BTMotion}{2.71}{10}{\textbf{2.71}} \\
\BTModel{deepseek}{DeepSeek-V4.1-Flash}{\cite{deepseekai2026deepseekv41flash}}
& \BTBar{BTExec}{2.00}{100}{2.00\%}
& \BTBar{BTExec}{88.00}{100}{88.00\%}
& \BTBar{BTVisual}{2.50}{10}{2.50}
& \BTBar{BTVisual}{2.08}{10}{2.08}
& \BTBar{BTMotion}{0.00}{10}{0.00}
& \BTBar{BTMotion}{0.11}{10}{0.11} \\
\BTModel{qwen}{Qwen3.8-Max}{\cite{alibaba2026qwen38max}}
& \BTBar{BTExec}{9.00}{100}{9.00\%}
& \BTBar{BTExec}{62.00}{100}{62.00\%}
& \BTBar{BTVisual}{0.67}{10}{0.67}
& \BTBar{BTVisual}{1.95}{10}{1.95}
& \BTBar{BTMotion}{0.00}{10}{0.00}
& \BTBar{BTMotion}{0.00}{10}{0.00} \\
\BTModel{kimi}{Kimi K3}{\cite{kimiteam2026kimik3}}
& \BTBar{BTExec}{10.00}{100}{10.00\%}
& \BTBar{BTExec}{79.00}{100}{79.00\%}
& \BTBar{BTVisual}{1.90}{10}{1.90}
& \BTBar{BTVisual}{2.05}{10}{2.05}
& \BTBar{BTMotion}{0.00}{10}{0.00}
& \BTBar{BTMotion}{0.00}{10}{0.00} \\
\BTModel{zai}{GLM-5.3-Flash}{\cite{zai2026glm53flash}}
& \BTBar{BTExec}{22.00}{100}{22.00\%}
& \BTBar{BTExec}{79.00}{100}{79.00\%}
& \BTBar{BTVisual}{2.50}{10}{2.50}
& \BTBar{BTVisual}{2.49}{10}{2.49}
& \BTBar{BTMotion}{0.33}{10}{0.33}
& \BTBar{BTMotion}{0.18}{10}{0.18} \\
\BTModel{claude}{Claude Fable 5.1}{\cite{anthropic2026fable51}}
& \BTBar{BTExec}{91.00}{100}{\textbf{91.00\%}}
    & \BTBar{BTExec}{93.00}{100}{\textbf{93.00\%}}
& \BTBar{BTVisual}{3.76}{10}{3.76}
& \BTBar{BTVisual}{3.75}{10}{3.75}
& \BTBar{BTMotion}{2.01}{10}{2.01}
& \BTBar{BTMotion}{1.98}{10}{1.98} \\
\bottomrule
\end{tabular}
\endgroup
\end{table}

\begin{table}[!htbp]
\centering
\caption{\textbf{Quantitative results on T3-1: Component-Constrained Synthesis.}}
\label{tab:app_t3-1}
\begingroup
\BTSetup{6}
\begin{tabular}{@{}>{\raggedright\arraybackslash}m{\BTmodelwidth}*{6}{>{\centering\arraybackslash}m{\BTbarwidth}}@{}}
\toprule
\multirow[c]{2}{\BTmodelwidth}[-\dimexpr(\aboverulesep+\belowrulesep+\cmidrulewidth)/2\relax]{\centering\textbf{Model}}
& \multicolumn{2}{c}{$Executability \uparrow$}
& \multicolumn{2}{c}{$Visual \uparrow$}
& \multicolumn{2}{c}{$\mathrm{ECM} \uparrow$} \\
\cmidrule(lr){2-3}\cmidrule(lr){4-5}\cmidrule(lr){6-7}
& $E_1$ & $E_2$
& $V_1$ & $V_2$
& $\mathrm{ECM}_{1}$ & $\mathrm{ECM}_{2}$ \\
\midrule
\BTModel{bytedance}{Doubao-Seed-2.1-Pro}{\cite{bytedanceseed2026seed21}}
& \BTBar{BTExec}{9.52}{100}{9.52\%}
& \BTBar{BTExec}{84.13}{100}{84.13\%}
& \BTBar{BTVisual}{3.50}{10}{3.50}
& \BTBar{BTVisual}{2.66}{10}{2.66}
& \BTBar{BTMotion}{0.51}{10}{0.51}
& \BTBar{BTMotion}{0.16}{10}{0.16} \\
\BTModel{openai}{GPT-6 Astra}{\cite{openai2026gpt6astra}}
& \BTBar{BTExec}{44.44}{100}{44.44\%}
& \BTBar{BTExec}{85.71}{100}{85.71\%}
& \BTBar{BTVisual}{5.00}{10}{\textbf{5.00}}
& \BTBar{BTVisual}{4.76}{10}{\textbf{4.76}}
& \BTBar{BTMotion}{4.12}{10}{\textbf{4.12}}
& \BTBar{BTMotion}{3.90}{10}{\textbf{3.90}} \\
\BTModel{deepseek}{DeepSeek-V4.1-Flash}{\cite{deepseekai2026deepseekv41flash}}
& \BTBar{BTExec}{4.76}{100}{4.76\%}
& \BTBar{BTExec}{87.30}{100}{87.30\%}
& \BTBar{BTVisual}{2.00}{10}{2.00}
& \BTBar{BTVisual}{2.42}{10}{2.42}
& \BTBar{BTMotion}{1.39}{10}{1.39}
& \BTBar{BTMotion}{0.33}{10}{0.33} \\
\BTModel{qwen}{Qwen3.8-Max}{\cite{alibaba2026qwen38max}}
& \BTBar{BTExec}{0.00}{100}{0.00\%}
& \BTBar{BTExec}{84.13}{100}{84.13\%}
& \BTBar{BTVisual}{0.00}{10}{0.00}
& \BTBar{BTVisual}{1.92}{10}{1.92}
& \BTBar{BTMotion}{0.00}{10}{0.00}
& \BTBar{BTMotion}{0.03}{10}{0.03} \\
\BTModel{kimi}{Kimi K3}{\cite{kimiteam2026kimik3}}
& \BTBar{BTExec}{6.35}{100}{6.35\%}
& \BTBar{BTExec}{95.24}{100}{95.24\%}
& \BTBar{BTVisual}{2.75}{10}{2.75}
& \BTBar{BTVisual}{2.48}{10}{2.48}
& \BTBar{BTMotion}{0.20}{10}{0.20}
& \BTBar{BTMotion}{0.42}{10}{0.42} \\
\BTModel{zai}{GLM-5.3-Flash}{\cite{zai2026glm53flash}}
& \BTBar{BTExec}{20.63}{100}{20.63\%}
& \BTBar{BTExec}{71.43}{100}{71.43\%}
& \BTBar{BTVisual}{3.23}{10}{3.23}
& \BTBar{BTVisual}{3.27}{10}{3.27}
& \BTBar{BTMotion}{0.50}{10}{0.50}
& \BTBar{BTMotion}{0.55}{10}{0.55} \\
\BTModel{claude}{Claude Fable 5.1}{\cite{anthropic2026fable51}}
& \BTBar{BTExec}{98.41}{100}{\textbf{98.41\%}}
& \BTBar{BTExec}{100.00}{100}{\textbf{100.00\%}}
& \BTBar{BTVisual}{4.39}{10}{4.39}
& \BTBar{BTVisual}{4.38}{10}{4.38}
& \BTBar{BTMotion}{2.77}{10}{2.77}
& \BTBar{BTMotion}{2.75}{10}{2.75} \\
\bottomrule
\end{tabular}
\endgroup
\end{table}

\begin{table}[!htbp]
\centering
\caption{\textbf{Quantitative results on T3-2: Goal-Constrained Synthesis.}}
\label{tab:app_t3-2}
\begingroup
\BTSetup{8}
\begin{tabular}{@{}>{\raggedright\arraybackslash}m{\BTmodelwidth}*{8}{>{\centering\arraybackslash}m{\BTbarwidth}}@{}}
\toprule
\multirow[c]{2}{\BTmodelwidth}[-\dimexpr(\aboverulesep+\belowrulesep+\cmidrulewidth)/2\relax]{\centering\textbf{Model}}
& \multicolumn{2}{c}{$Executability \uparrow$}
& \multicolumn{2}{c}{$Visual \uparrow$}
& \multicolumn{2}{c}{$\mathrm{ECM} \uparrow$}
& \multicolumn{2}{c}{Goal Satisfaction $\uparrow$} \\
\cmidrule(lr){2-3}\cmidrule(lr){4-5}\cmidrule(lr){6-7}\cmidrule(lr){8-9}
& $E_1$ & $E_2$
& $V_1$ & $V_2$
& $\mathrm{ECM}_{1}$ & $\mathrm{ECM}_{2}$
& $\mathbf{G}_{1}$ & $\mathbf{G}_{2}$ \\
\midrule
\BTModel{bytedance}{Doubao-Seed-2.1-Pro}{\cite{bytedanceseed2026seed21}}
& \BTBar{BTExec}{1.67}{100}{1.67\%}
& \BTBar{BTExec}{70.00}{100}{70.00\%}
& \BTBar{BTVisual}{3.00}{10}{3.00}
& \BTBar{BTVisual}{2.60}{10}{2.60}
& \BTBar{BTMotion}{1.00}{10}{1.00}
& \BTBar{BTMotion}{0.48}{10}{0.48}
& \BTBar{BTGoal}{15.38}{100}{15.38\%}
& \BTBar{BTGoal}{10.68}{100}{10.68\%} \\
\BTModel{openai}{GPT-6 Astra}{\cite{openai2026gpt6astra}}
& \BTBar{BTExec}{40.00}{100}{40.00\%}
& \BTBar{BTExec}{91.67}{100}{\textbf{91.67\%}}
& \BTBar{BTVisual}{4.92}{10}{\textbf{4.92}}
& \BTBar{BTVisual}{4.73}{10}{\textbf{4.73}}
& \BTBar{BTMotion}{2.88}{10}{\textbf{2.88}}
& \BTBar{BTMotion}{2.78}{10}{\textbf{2.78}}
& \BTBar{BTGoal}{47.93}{100}{\textbf{47.93\%}}
& \BTBar{BTGoal}{44.81}{100}{\textbf{44.81\%}} \\
\BTModel{deepseek}{DeepSeek-V4.1-Flash}{\cite{deepseekai2026deepseekv41flash}}
& \BTBar{BTExec}{10.00}{100}{10.00\%}
& \BTBar{BTExec}{81.67}{100}{81.67\%}
& \BTBar{BTVisual}{2.50}{10}{2.50}
& \BTBar{BTVisual}{2.16}{10}{2.16}
& \BTBar{BTMotion}{0.50}{10}{0.50}
& \BTBar{BTMotion}{0.63}{10}{0.63}
& \BTBar{BTGoal}{11.45}{100}{11.45\%}
& \BTBar{BTGoal}{9.20}{100}{9.20\%} \\
\BTModel{qwen}{Qwen3.8-Max}{\cite{alibaba2026qwen38max}}
& \BTBar{BTExec}{0.00}{100}{0.00\%}
& \BTBar{BTExec}{81.67}{100}{81.67\%}
& \BTBar{BTVisual}{0.00}{10}{0.00}
& \BTBar{BTVisual}{2.24}{10}{2.24}
& \BTBar{BTMotion}{0.00}{10}{0.00}
& \BTBar{BTMotion}{0.45}{10}{0.45}
& \BTBar{BTGoal}{0.00}{100}{0.00\%}
& \BTBar{BTGoal}{7.16}{100}{7.16\%} \\
\BTModel{kimi}{Kimi K3}{\cite{kimiteam2026kimik3}}
& \BTBar{BTExec}{3.33}{100}{3.33\%}
& \BTBar{BTExec}{81.67}{100}{81.67\%}
& \BTBar{BTVisual}{4.00}{10}{4.00}
& \BTBar{BTVisual}{3.31}{10}{3.31}
& \BTBar{BTMotion}{1.50}{10}{1.50}
& \BTBar{BTMotion}{0.78}{10}{0.78}
& \BTBar{BTGoal}{45.93}{100}{45.93\%}
& \BTBar{BTGoal}{22.20}{100}{22.20\%} \\
\BTModel{zai}{GLM-5.3-Flash}{\cite{zai2026glm53flash}}
& \BTBar{BTExec}{10.00}{100}{10.00\%}
& \BTBar{BTExec}{30.00}{100}{30.00\%}
& \BTBar{BTVisual}{3.17}{10}{3.17}
& \BTBar{BTVisual}{3.56}{10}{3.56}
& \BTBar{BTMotion}{0.00}{10}{0.00}
& \BTBar{BTMotion}{0.50}{10}{0.50}
& \BTBar{BTGoal}{8.61}{100}{8.61\%}
& \BTBar{BTGoal}{17.64}{100}{17.64\%} \\
\BTModel{claude}{Claude Fable 5.1}{\cite{anthropic2026fable51}}
& \BTBar{BTExec}{73.33}{100}{\textbf{73.33\%}}
& \BTBar{BTExec}{78.33}{100}{78.33\%}
& \BTBar{BTVisual}{4.39}{10}{4.39}
& \BTBar{BTVisual}{4.43}{10}{4.43}
& \BTBar{BTMotion}{2.30}{10}{2.30}
& \BTBar{BTMotion}{2.36}{10}{2.36}
& \BTBar{BTGoal}{39.32}{100}{39.32\%}
& \BTBar{BTGoal}{39.97}{100}{39.97\%} \\
\bottomrule
\end{tabular}
\endgroup
\end{table}

\subsection{Result Visualizations}
We present one representative evaluation result for T1, T2, T3-1 and T3-2, showing the best-performing harness-based model together with API-based models spanning highest, medium, and lowest performance levels.

\begin{figure}[tbp]
    \centering
    \includegraphics[width=0.9\textwidth]{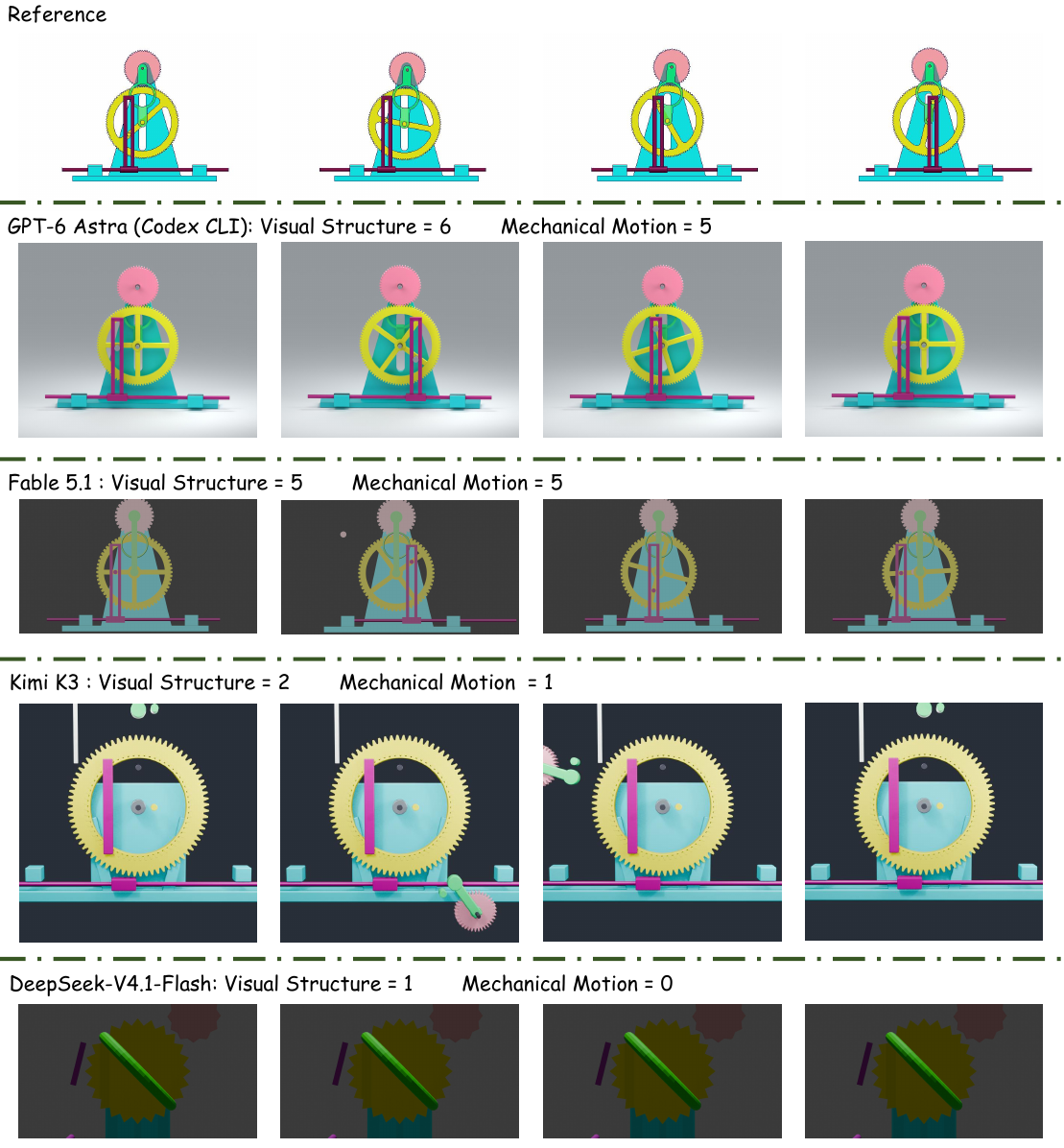}
    \caption{\textbf{Representative qualitative results on T1: Mechanism Executability.}
    Examples from the best-performing harness-based model and API-based models with high, medium, and low performance.}
    \label{fig:app_t1}
\end{figure}

Representative T1 results are shown in Fig.~\ref{fig:app_t1}.

\begin{figure}[tbp]
    \centering
    \includegraphics[width=0.9\textwidth]{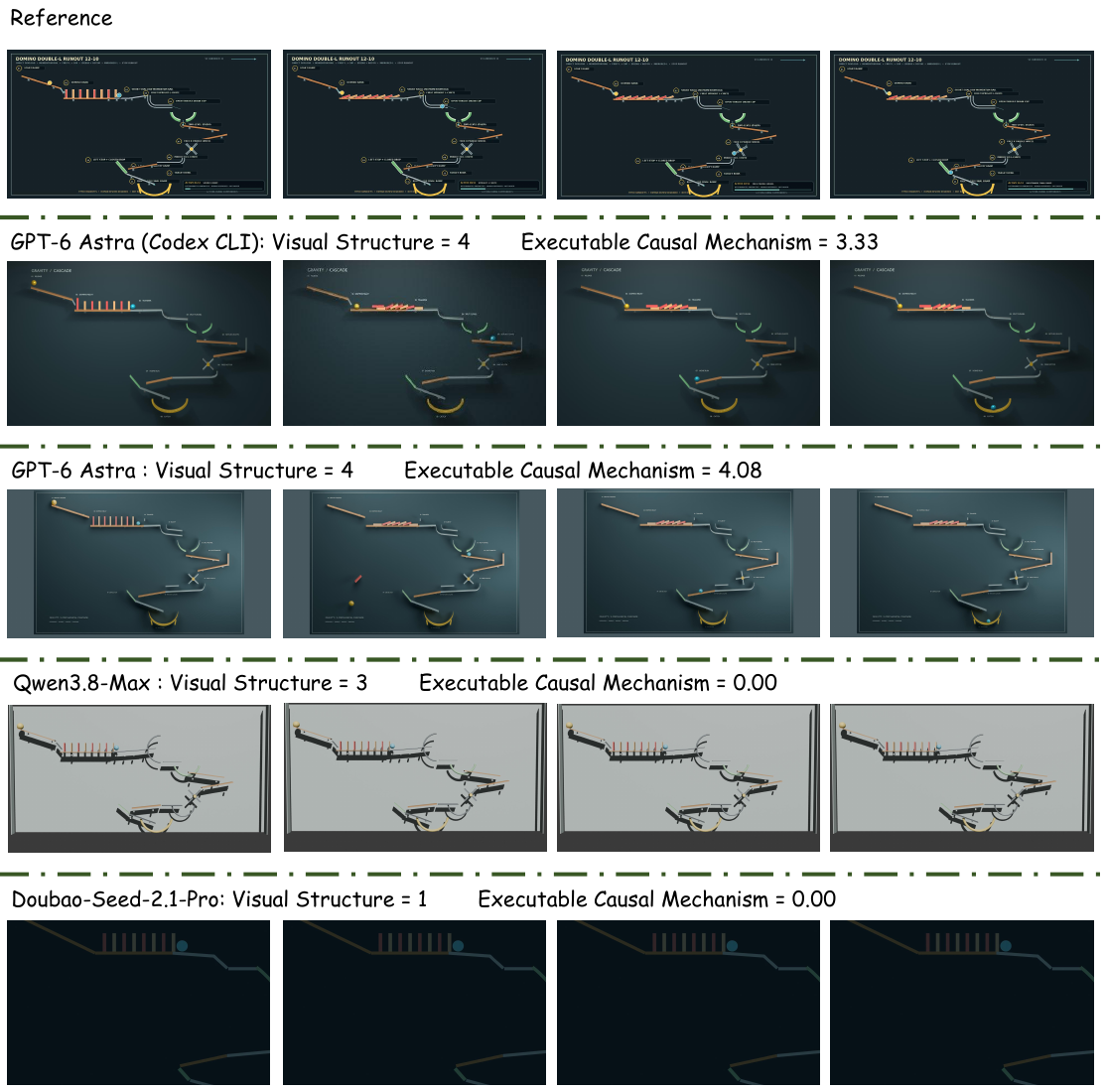}
    \caption{\textbf{Representative qualitative results on T2: Compositional Executability.}
    Examples from the best-performing harness-based model and API-based models with high, medium, and low performance.}
    \label{fig:app_t2}
\end{figure}

Representative T2 results are shown in Fig.~\ref{fig:app_t2}.

\begin{figure}[tbp]
    \centering
    \includegraphics[width=0.9\textwidth]{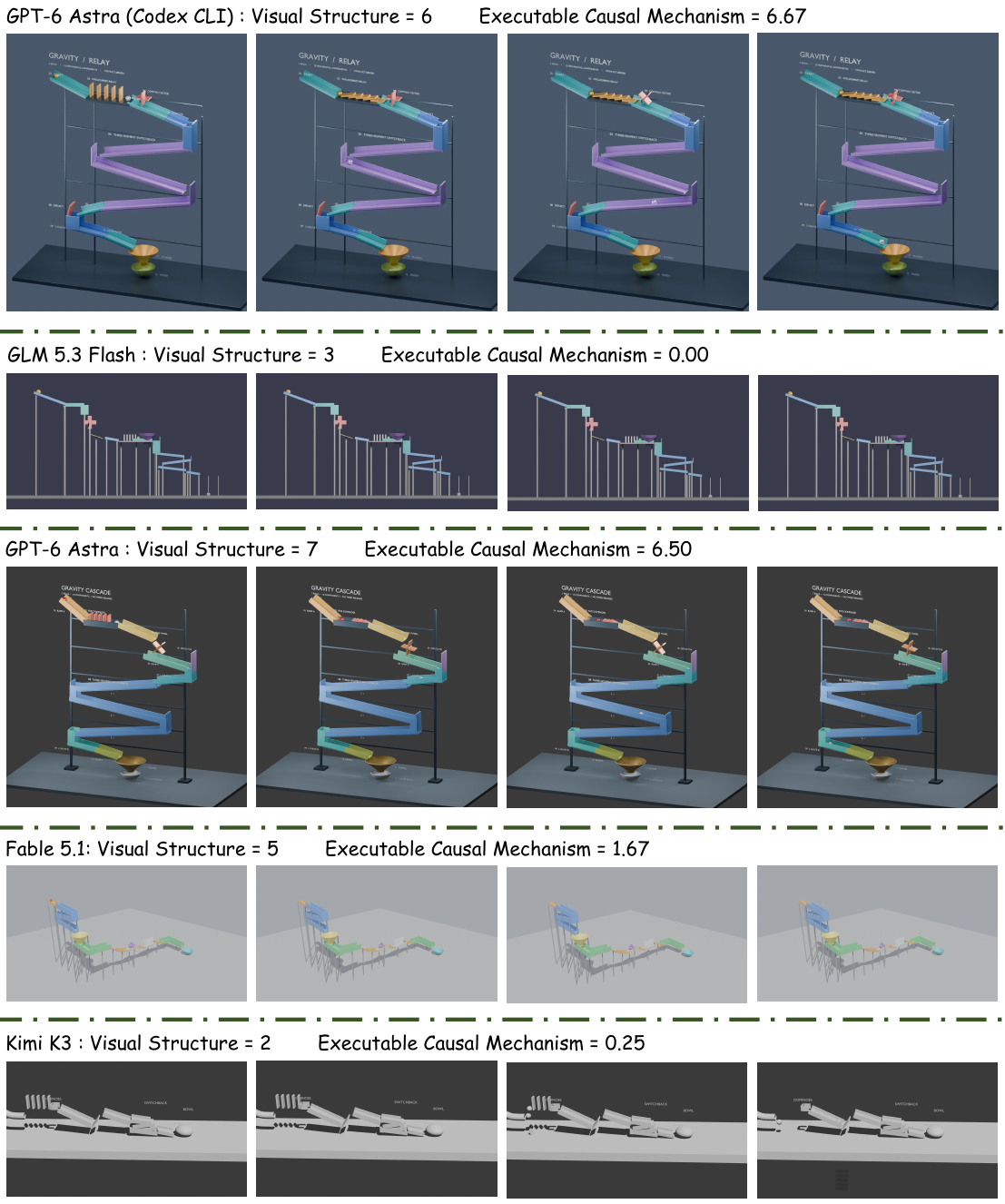}
    \caption{\textbf{Representative qualitative results on T3-1: Component-Constrained Synthesis.}
    Examples from the best-performing harness-based model and API-based models with high, medium, and low performance.}
    \label{fig:app_t3_1}
\end{figure}

\begin{figure}[tbp]
    \centering
    \includegraphics[width=0.9\textwidth]{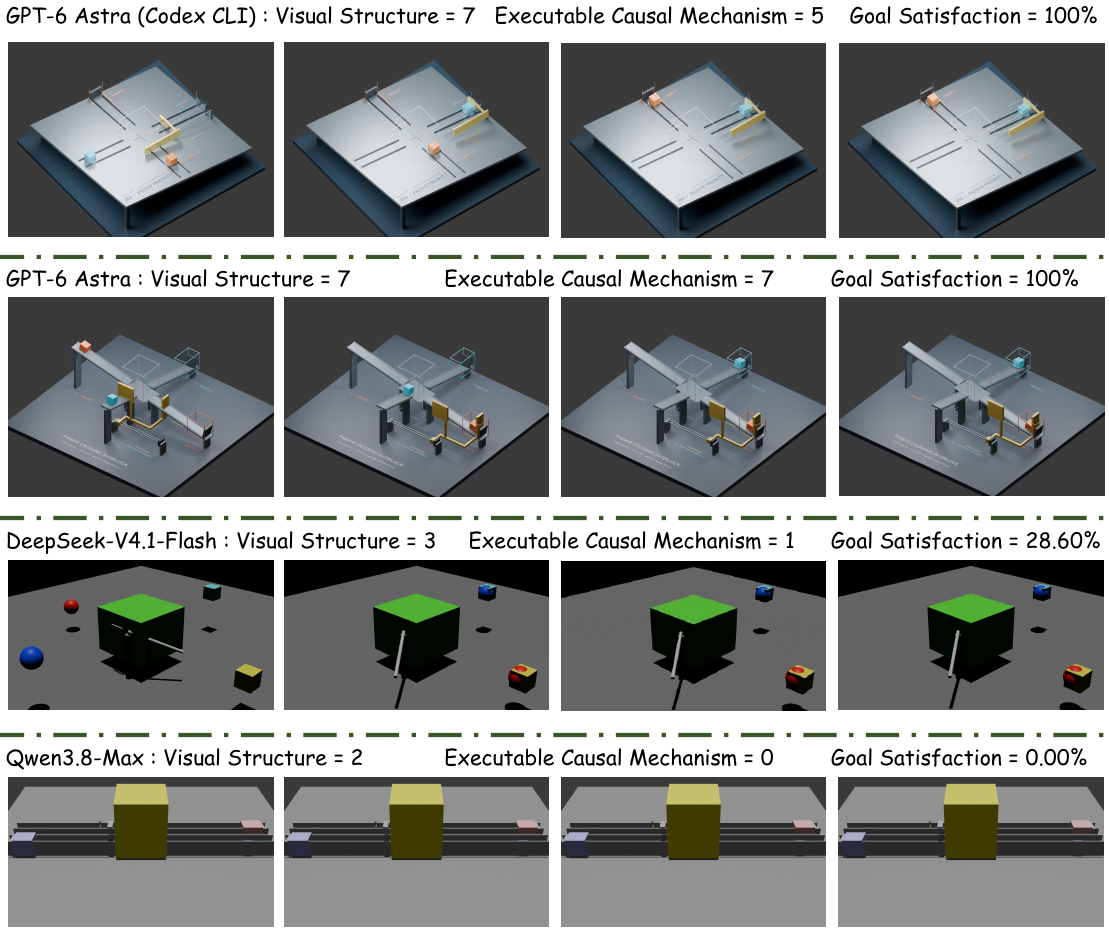}
    \caption{\textbf{Representative qualitative results on T3-2: Goal-Constrained Synthesis.}
    Examples from the best-performing harness-based model and API-based models with high, medium, and low performance.}
    \label{fig:app_t3_2}
\end{figure}

\clearpage
As for T3-1, the results are shown in Fig.~\ref{fig:app_t3_1}, and the task specification is shown below:

\begin{lstlisting}[
  basicstyle=\ttfamily\scriptsize,
  breaklines=true,
  breakatwhitespace=false,
  columns=fullflexible,
  keepspaces=true
]
{"components": [
  {"name": "Three-Segment Switchback Track",
    "definition": "Three track segments are arranged sequentially at different heights, with adjacent segments extending in alternating opposite directions."},
  {"name": "Ramp",
    "definition": "An inclined board with a continuous surface."},
  {"name": "Catch Bowl",
    "definition": "A bowl-shaped container with an upward-facing opening and a concave inner surface."},
  {"name": "Ramp",
    "definition": "An inclined board with a continuous surface."},
  {"name": "Deflector Plate",
    "definition": "A fixed rigid plate positioned near a track, chute, or motion path."},
  {"name": "L-Shaped Chute",
    "definition": "A continuous chute composed of two channel sections oriented in different directions, forming an approximately L-shaped profile."},
  {"name": "Five-Domino Chain",
    "definition": "Five upright dominoes are arranged sequentially with small gaps between adjacent dominoes."},
  {"name": "L-Shaped Chute",
    "definition": "A continuous chute composed of two channel sections oriented in different directions, forming an approximately L-shaped profile."},
  {"name": "Ramp",
    "definition": "An inclined board with a continuous surface."},
  {"name": "Four-Paddle Wheel",
    "definition": "Four paddles are arranged around a central axle and can rotate about the axle when contacted by a moving object."},
  {"name": "Funnel",
    "definition": "A hollow container with a wide upper opening that narrows toward a smaller lower section or outlet."},
  {"name": "Ramp",
    "definition": "An inclined board with a continuous surface."}
],
  "component_count": 12,
  "ball_count": 2}
\end{lstlisting}

As for T3-2, the results are shown in Fig.~\ref{fig:app_t3_2}, and the task specification is shown below:
\begin{lstlisting}[
  basicstyle=\ttfamily\scriptsize,
  breaklines=true,
  breakatwhitespace=false,
  columns=fullflexible,
  keepspaces=true
]
{
  "Goal": "Start payload_B's approach before payload_A enters shared_region, but mechanically give payload_A priority through the region.",
  "Required Objects": {
    "ground_plane": "Static physical support for the mechanism.",
    "payload_A": "Transported body assigned exclusively to target_A.",
    "target_A": "Fixed, visibly marked receiving volume for payload_A; initially empty.",
    "payload_B": "Transported body assigned exclusively to target_B.",
    "target_B": "Fixed, visibly marked receiving volume for payload_B; initially empty.",
    "checkpoint_A": "Directed plane strictly inside shared_region on payload_A's path.",
    "checkpoint_B": "Directed plane strictly inside shared_region on payload_B's path.",
    "shared_region": "Fixed visibly outlined crossing volume used by all payload routes; its boundary must be declared before simulation.",
    "object_A": "Task-relevant passive mechanical component; shape and linkage are freely designed."
  },
  "State Definitions": {
    "state_A0": "object_A's declared initial physical configuration.",
    "state_A1": "object_A's distinct physical configuration 1; its measurable pose region is declared before simulation and disjoint from the other declared configurations."
  },
  "Event Definitions": {
    "E_change_A1": "First entry of object_A into state_A1.",
    "E_cross_A_A": "First forward crossing of checkpoint_A by the center of payload_A through its declared aperture.",
    "E_cross_B_B": "First forward crossing of checkpoint_B by the center of payload_B through its declared aperture.",
    "E_enter_A": "First overlap of any collision geometry of payload_A with shared_region.",
    "E_enter_B": "First overlap of any collision geometry of payload_B with shared_region.",
    "E_exit_A": "First full clearance of payload_A's collision geometry from shared_region after E_enter_A.",
    "E_exit_B": "First full clearance of payload_B's collision geometry from shared_region after E_enter_B.",
    "E_start_B": "First sustained departure of payload_B from its initial pose beyond the motion threshold.",
    "E_target_A": "First full containment of payload_A's collision geometry in target_A.",
    "E_target_B": "First full containment of payload_B's collision geometry in target_B."
  },
  "Initial State": {
    "I01": "payload_A is stationary at simulation start.",
    "I02": "payload_A is outside target_A at simulation start.",
    "I03": "payload_B is stationary at simulation start.",
    "I04": "payload_B is outside target_B at simulation start.",
    "I05": "payload_A is fully outside shared_region at simulation start.",
    "I06": "payload_B is fully outside shared_region at simulation start.",
    "I07": "object_A occupies state_A0 at simulation start."
  },
  "Rules and Required Physical Dependencies": {
    "R01": "E_enter_A and E_enter_B both occur, with E_enter_A strictly before E_enter_B.",
    "R02": "payload_B remains fully outside shared_region until E_exit_A.",
    "R03": "E_enter_A and E_cross_A_A both occur, with E_enter_A strictly before E_cross_A_A. E_cross_A_A and E_exit_A both occur, with E_cross_A_A strictly before E_exit_A.",
    "R04": "E_enter_A and E_exit_A both occur, with E_enter_A strictly before E_exit_A.",
    "R05": "E_exit_A and E_change_A1 both occur, with E_change_A1 no earlier than E_exit_A; simultaneous events within one simulation step are allowed. E_change_A1 and E_enter_B both occur, with E_enter_B no earlier than E_change_A1; simultaneous events within one simulation step are allowed. Clearance of payload_A must physically change the entry-control mechanism object_A. object_A must mechanically permit the already-moving payload_B to enter shared_region.",
    "R06": "E_exit_A and E_enter_B both occur, with E_exit_A strictly before E_enter_B.",
    "R07": "payload_A and payload_B never physically contact one another.",
    "R08": "E_exit_A and E_target_A both occur, with E_exit_A strictly before E_target_A.",
    "R09": "E_enter_B and E_cross_B_B both occur, with E_enter_B strictly before E_cross_B_B. E_cross_B_B and E_exit_B both occur, with E_cross_B_B strictly before E_exit_B.",
    "R10": "E_exit_B and E_target_B both occur, with E_exit_B strictly before E_target_B.",
    "R11": "From E_change_A1 through the end of the final hold, object_A remains in state_A1.",
    "R12": "payload_A and payload_B never simultaneously overlap shared_region. payload_A never re-enters shared_region after E_exit_A. payload_B never re-enters shared_region after E_exit_B.",
    "R13": "E_start_B and E_enter_A both occur, with E_start_B strictly before E_enter_A."
  },
  "Required Final State": {
    "F01": "payload_A remains fully inside target_A throughout the final hold interval.",
    "F02": "payload_B remains fully inside target_B throughout the final hold interval.",
    "F03": "object_A remains in state_A1 throughout the final hold interval."
  }
}
\end{lstlisting}

\end{document}

%% file: table/benchmark-table-style.tex
\usepackage{graphicx}
\usepackage{booktabs}
\usepackage{array}
\usepackage{multirow}
\usepackage{xcolor}
\usepackage{tikz}
\usepackage{etoolbox}

\definecolor{BTExec}{HTML}{A998ED}
\definecolor{BTVisual}{HTML}{EDB75D}
\definecolor{BTMotion}{HTML}{68AEA5}
\definecolor{BTGoal}{HTML}{A998ED}

\newlength{\BTmodelwidth}
\newlength{\BTbarwidth}
\newcommand{\BTiconpath}{model-icons}

\newcommand{\BTSetup}[1]{%
  \fontsize{8.5pt}{10.5pt}\selectfont
  \setlength{\tabcolsep}{2pt}%
  \renewcommand{\arraystretch}{1.10}%
  \setlength{\extrarowheight}{2pt}%
  \ifnum#1=8
    \setlength{\BTmodelwidth}{0.32\linewidth}%
  \else
    \setlength{\BTmodelwidth}{0.36\linewidth}%
  \fi
  \setlength{\BTbarwidth}{%
    \dimexpr(\linewidth-\BTmodelwidth-\tabcolsep*2*#1)/#1\relax}%
}

\newcommand{\BTIcon}[1]{%
  \IfFileExists{\BTiconpath/#1.png}{%
    \raisebox{-0.18em}{%
      \includegraphics[width=1.15em,height=1.15em,keepaspectratio]{\BTiconpath/#1.png}}%
  }{%
    \PackageWarning{benchmark-tables}{Missing icon: \BTiconpath/#1.png}%
    \makebox[1.15em]{\textbullet}%
  }%
}

\newcommand{\BTModel}[3]{%
  \BTIcon{#1}\hspace{0.35em}#2\par
  {\fontsize{6.5pt}{8pt}\selectfont #3\par}%
}

\newcommand{\BTBar}[4]{%
  \ifstrequal{#2}{-}{%
    \makebox[\BTbarwidth][c]{-}%
  }{%
    \begingroup
    \pgfmathsetmacro{\BTfraction}{max(0,min(1,(#2)/(#3)))}%
    \begin{tikzpicture}[x=\BTbarwidth,y=14pt,baseline=(BTvalue.base)]
      \path[use as bounding box] (0,0) rectangle (1,1);
      \begin{scope}
        \clip[rounded corners=1.2pt] (0,0) rectangle (1,1);
        \fill[#1!15!white] (0,0) rectangle (1,1);
        \ifdim\BTfraction pt>0pt
          \shade[left color=#1!85!white,right color=#1!48!white]
            (0,0) rectangle (\BTfraction,1);
        \fi
      \end{scope}
      \node[inner sep=0pt,outer sep=0pt,
        font=\normalfont\fontsize{8pt}{9pt}\selectfont]
        (BTvalue) at (0.5,0.5) {#4};
    \end{tikzpicture}%
    \endgroup
  }%
}

%% file: math_commands.tex
\usepackage{amsmath,amsfonts,bm}

\def\eqref#1{equation~\ref{#1}}
\def\1{\bm{1}}

\DeclareMathAlphabet{\mathsfit}{\encodingdefault}{\sfdefault}{m}{sl}
\SetMathAlphabet{\mathsfit}{bold}{\encodingdefault}{\sfdefault}{bx}{n}